%% file: main_arXiv.tex
\documentclass[letterpaper]{article} 
\usepackage{aaai2026}  
\usepackage{times}  
\usepackage{helvet}  
\usepackage{courier}  
\usepackage[hyphens]{url}  
\usepackage{graphicx} 
\usepackage{natbib}  
\usepackage{caption} 
\usepackage{algorithm}
\usepackage{algorithmic}
\usepackage{enumitem}
\usepackage{multirow}
\usepackage{booktabs}
\usepackage{pifont} 
\usepackage{subcaption} 
\usepackage{comment}
\usepackage{arydshln}

\usepackage{newfloat}
\usepackage{listings}
\DeclareCaptionStyle{ruled}{labelfont=normalfont,labelsep=colon,strut=off} 
\floatstyle{ruled}
\newfloat{listing}{tb}{lst}{}
\floatname{listing}{Listing}
\newcommand{\SimpleQA}{\texttt{Video} \texttt{SimpleQA} }
\usepackage{xcolor}
\definecolor{my_green}{RGB}{51,102,0}
\definecolor{my_red}{RGB}{204, 0, 0}
\definecolor{firstBest}{rgb}{0.86, 1, 0.86} 
\newcommand{\cmark}{{\ding{51}}} 
\newcommand{\xmark}{{\ding{55}}} 
\usepackage{xspace}
\makeatletter
\DeclareRobustCommand\onedot{\futurelet\@let@token\@onedot}
\def\@onedot{\ifx\@let@token.\else.\null\fi\xspace}
\def\eg{\emph{e.g}\onedot} 
\def\ie{\emph{i.e}\onedot} 
\def\cf{\emph{c.f}\onedot} 
\def\etc{\emph{etc}\onedot} \def\vs{\emph{vs}\onedot}

\makeatother

\title{Video SimpleQA: Towards Factuality Evaluation in \\Large Video Language Models}
\author{
    Meng Cao\textsuperscript{\rm 1,2}\footnotemark[1],
    Pengfei Hu\textsuperscript{\rm 1,2}\footnotemark[1], 
    Yingyao Wang\textsuperscript{\rm 2},
    Jihao Gu\textsuperscript{\rm 2},
    Haoran Tang\textsuperscript{\rm 1},\\
    Haoze Zhao\textsuperscript{\rm 1},
    Chen Wang\textsuperscript{\rm 2},
    Jiahua Dong\textsuperscript{\rm 2},
    Wangbo Yu\textsuperscript{\rm 3},
    Ge Zhang\textsuperscript{\rm 4},
    Jun Song\textsuperscript{\rm 2},\\
    Xiang Li\textsuperscript{\rm 2},
    Bo Zheng\textsuperscript{\rm 2},
    Ian Reid\textsuperscript{\rm 1},
    Xiaodan Liang\textsuperscript{\rm 1,5}$^{\dagger}$\\
    \textcolor{blue}{\url{https://videosimpleqa.github.io/}}\\
}
\affiliations{
    \textsuperscript{\rm 1}MBZUAI
    \textsuperscript{\rm 2}Taobao\&Tmall Group
    \textsuperscript{\rm 3}Peking University
    \textsuperscript{\rm 4}MAP
    \textsuperscript{\rm 5}Sun Yat-sen University \\
    \small{$^*$Equal contribution. $^\dag$Corresponding author.}

}

\usepackage{bibentry}

\begin{document}




\input{table_figs/figTeaserTop}
\input{sec/0_abstract}    
\input{sec/1_intro}

\input{sec/2_related}
\input{sec/3_method}
\input{sec/4_experiments}
\input{sec/5_con}
\input{sec/6_appendix}
    

\bibliography{aaai2026}

\end{document}

%% file: table_figs/figTeaserTop.tex
\twocolumn[{
	\renewcommand\twocolumn[1][]{#1}
	\maketitle
	\begin{center}
           \vspace{-2mm}
            \includegraphics[width=0.97\linewidth]{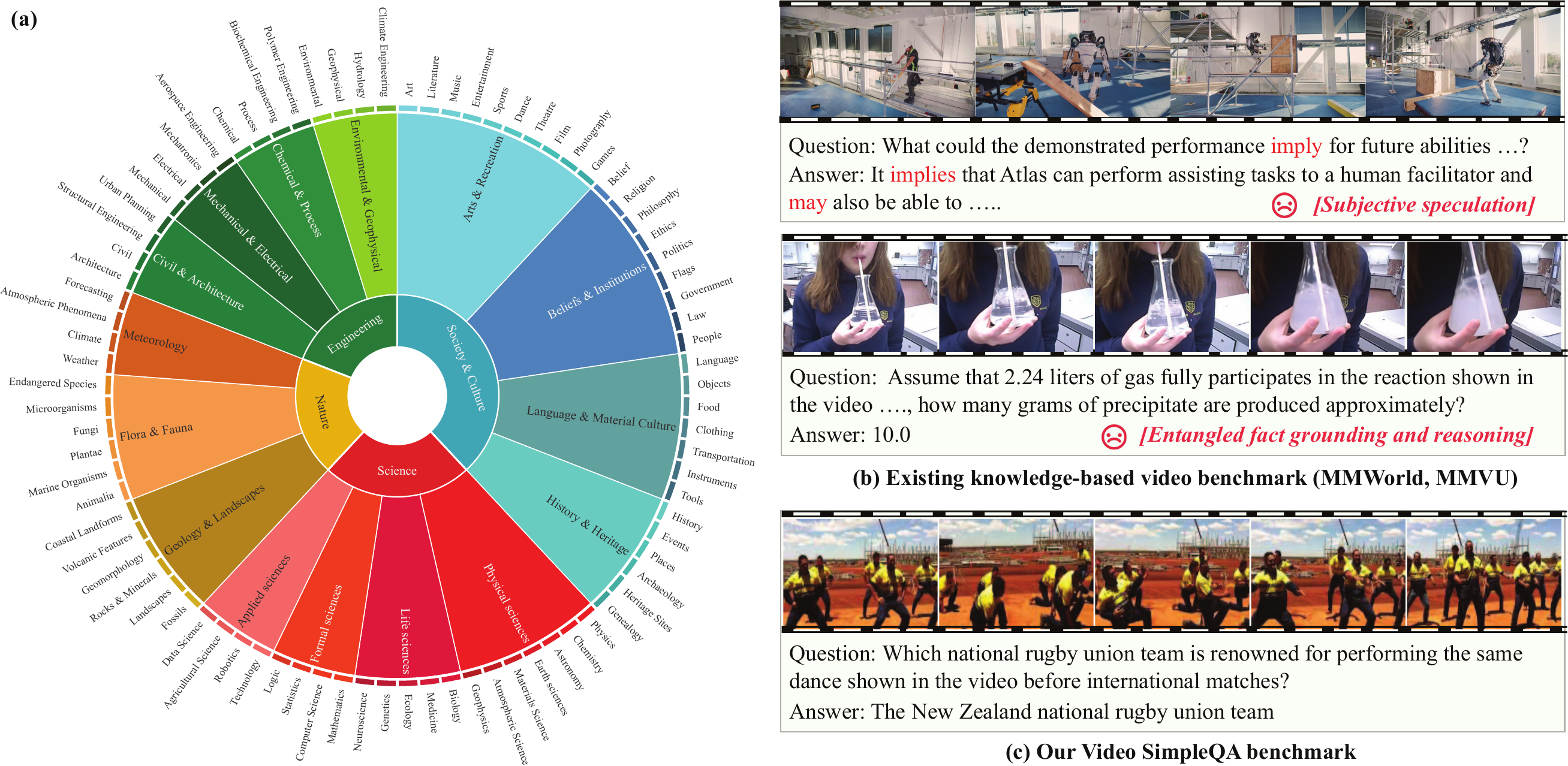}
		\captionsetup{type=figure}
	   \caption{(a) \textbf{The taxonomy} of \SimpleQA benchmark; (b) \textbf{Illustrations of existing knowledge-based video benchmarks} \cite{zhao2025mmvu,he2024mmworld} which may contain subjective speculation or conflate factual grounding with reasoning skills (\ie, mathematical calculation); (c) \textbf{Illustrations of our Video SimpleQA benchmark} with the fact-seeking question and definitive \& short-form answer with multi-hop external facts verified.}
		\label{fig:teaserTop}
	\end{center}
}]

%% file: sec/0_abstract.tex
\input{table_figs/figVis}
\begin{abstract}
Recent advancements in Large Video Language Models (LVLMs) have highlighted their potential for multi-modal understanding, yet evaluating their factual grounding in videos remains a critical unsolved challenge. To address this gap, we introduce \texttt{Video} \texttt{SimpleQA}, the first comprehensive benchmark tailored for factuality evaluation in video contexts. Our work differs from existing video benchmarks through the following key features: 1) \textbf{Knowledge required}:  demanding integration of external knowledge beyond the video’s explicit narrative; 2) \textbf{Multi-hop fact-seeking question}: Each question involves multiple explicit facts and requires strict factual grounding without hypothetical or subjective inferences. We also include per-hop single-fact-based sub-QAs alongside final QAs to enable fine-grained, step-by-step evaluation; 3) \textbf{Short-form definitive answer}: Answers are crafted as unambiguous and definitively correct in a short format with minimal scoring variance; 4) \textbf{Temporal grounded required}: Requiring answers to rely on one or more temporal segments in videos, rather than single frames. We extensively evaluate 33 state-of-the-art LVLMs and summarize key findings as follows: 1) Current LVLMs exhibit notable deficiencies in factual adherence, with the best-performing model o3 merely achieving an F-score of 66.3\%; 2) Most LVLMs are overconfident in what they generate, with self-stated confidence exceeding actual accuracy; 3) Retrieval-augmented generation demonstrates consistent improvements at the cost of additional inference time overhead; 4) Multi-hop QA demonstrates substantially degraded performance compared to single-hop sub-QAs, with first-hop object/event recognition emerging as the primary bottleneck. We position \SimpleQA as the cornerstone benchmark for video factuality assessment, aiming to steer LVLM development toward verifiable grounding in real-world contexts.
\end{abstract}


%% file: table_figs/figVis.tex
\begin{figure*}[ht]
	\centering
        \includegraphics[width=0.96\textwidth]{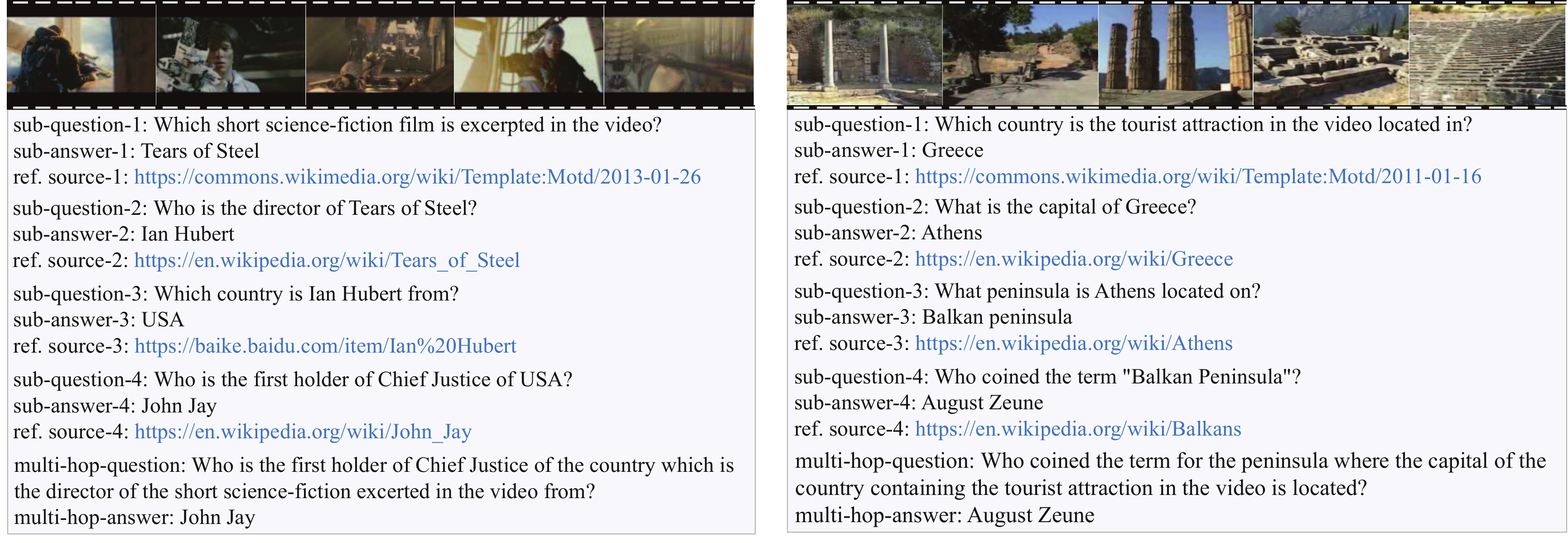}
	\caption{\textbf{Four-hop examples} in \SimpleQA including the final multi-hop QA and the decomposed per-fact sub-QAs.}
	\label{fig:vis}
\end{figure*}

%% file: sec/1_intro.tex
\section{Introduction} \label{sec:intro}

The substantial advancements in Large Language Models (LLMs) \cite{achiam2023gpt,reid2024gemini,touvron2023llama2} over the past few years have inaugurated a new frontier in artificial intelligence. Despite their remarkable capabilities, the factuality concern \cite{wang2024factuality,akhtar2023multimodal,wang2023survey} remains a critical challenge, \ie, how to ensure that the generated contents are consistent with factual knowledge and grounded in credible sources.\footnote{Please refer to \cite{wang2023survey,wang2024factuality} for the differentiation between the \emph{factuality} and the similar \emph{hallucination} concepts.} 

Existing research has primarily focused on evaluating factuality in text-based \cite{yu2022generate,pan2024unifying,lin2022truthfulqa,chern2023factool,gou2023critic} and image-based \cite{marino2019ok,wang2015explicit,wang2017fvqa,zellers2019recognition,jain2021select} scenarios. Recently, the SimpleQA benchmark \cite{wei2024measuring} introduced by OpenAI and its subsequent works \cite{he2024chinese,gu2025see,cheng2025simplevqa} streamline the factuality evaluation by considering only concise and fact-seeking questions, which enables standardized and tractable assessments. However, extending this paradigm to video contexts is under-explored and presents unique challenges due to the inherent temporal dynamics and procedural knowledge. 
\input{table_figs/tabBenchmarkCompare}
To bridge this gap, we present \texttt{Video} \texttt{SimpleQA}, a comprehensive factuality evaluation benchmark tailored for Large Video Language Models (LVLMs). As shown in Figure \ref{fig:teaserTop} and Figure \ref{fig:vis}, \SimpleQA is composed of multi-hop fact-seeking questions and short-form definitive answers. Compared to previous video benchmarks, \SimpleQA stands out with the following advancements:
\begin{itemize}[topsep=2pt, partopsep=0pt, leftmargin=13pt, parsep=0pt, itemsep=3pt]
    \item \textbf{Knowledge required}: Beyond comprehending the visual content, \SimpleQA necessitates the integration of external knowledge that is not explicitly presented in the video narrative, \eg, domain-specific information, contextual background, commonsense. 

    \item \textbf{Multi-hop fact-seeking question}: Questions necessitate strict adherence to factual grounding principles, \emph{eliminating any hypothetical or subjective inferences}. In addition, each question is constructed to involve \emph{multiple explicitly identifiable facts}. To achieve this, beyond the final multi-hop question-answer (QA) pairs, we additionally provide per-fact specific sub-QA annotations, which facilitate fine-grained evaluation of model performance at each fact-grounding hop and help pinpoint exactly which hop fails in factual grounding (\cf Figure \ref{fig:vis}).
        
    \item \textbf{Short-form definitive answer}: All the answers are unambiguous, universally agreed upon, consistent over time, and invariant to individual perspectives. Following SimpleQA \cite{wei2024measuring}, the answers also advocate the short-form paradigm, which establishes reliable factual assessment with low run-to-run variance.

    \item \textbf{Temporal grounded:} Answering questions in \SimpleQA should refer to one or more temporal segments in the video, rather than relying on a single frame.
    
        
    

\end{itemize}
While existing knowledge-based \cite{garcia2020knowit,zhang2024worldqa,hu2025video} and recent discipline-based \cite{zhao2025mmvu,he2024mmworld} benchmarks may appear similar to our \SimpleQA, our benchmark features several distinct characteristics (\cf Table \ref{tab:benchComparison}):
\begin{itemize}[topsep=2pt, partopsep=0pt, leftmargin=13pt, parsep=0pt, itemsep=3pt]
    \item  \textbf{Open-domain}: While KnowIT-VQA \cite{garcia2020knowit} is constrained to TV shows, and MMVU \cite{zhao2025mmvu} as well as MMWorld \cite{he2024mmworld} focus on discipline-specific knowledge, our \SimpleQA encompasses open-domain video types and questions.

    \item \textbf{Objective QA}: Our benchmark is explicitly designed for factuality evaluation through \textbf{objective} factual assertions, in contrast to existing benchmarks that often involve varying degrees of subjectivity, even those focusing on disciplinary knowledge. For instance, as shown in Figure 1(b) top, MMWorld \cite{he2024mmworld} includes cases requiring predictions about a robot’s future capabilities—introducing subjective speculation and personal judgment, which deviates from our goal of evidence-based and objective evaluation.\footnote{More examples are available in the supplementary material.\label{supple}}

    \item \textbf{Factuality exclusive}: Discipline-based benchmarks often conflate external knowledge retrieval with reasoning skills (\eg, numerical calculations). For example, the case in Figure 1(b) bottom from MMVU \cite{zhao2025mmvu} requires LVLMs to both recognize a chemical reaction in the video and perform numeric computations based on the question context. This \emph{coupling} makes it difficult to pinpoint the error source—whether due to incorrect fact identification (\eg, failing to detect the reaction) or faulty reasoning (\eg, miscalculating). In contrast, \SimpleQA \textbf{exclusively} focuses on fact identification, providing a clearer assessment of LVLMs’ fact-grounding ability.\footref{supple}

   \item \textbf{Multi-hop fact decomposition}: As shown in Figure~\ref{fig:vis}, \SimpleQA includes not only the final multi-hop QA pairs but also the decomposed \emph{per-fact sub-QAs}, enabling fine-grained evaluations. While some cases in MMVU \cite{zhao2025mmvu} also involve knowledge from multiple external sources, they do not provide such explicit per-fact decomposition, making it difficult to assess how each individual fact contributes to the final answer.
\end{itemize}

We conduct comprehensive evaluations of 33 state-of-the-art LVLMs on \texttt{Video} \texttt{SimpleQA}, revealing several critical insights: 1) \textbf{Significant performance gap}: Both proprietary and open-source LVLMs substantially underperform compared to human expertise; 2) \textbf{Overconfidence bias}: Most LVLMs exhibit systematic overconfidence in their predictions despite output inaccuracies, with notable variations in calibration quality (\cf Figure \ref{fig:calibration}); 3) \textbf{Efficiency-performance tradeoff}: Retrieval-Augmented Generation (RAG) yields significant gains at the cost of inference efficiency (\cf Table \ref{tab:rag}); 4) \textbf{Multi-hop performance bottleneck}: Multi-hop QA performance significantly lags behind single-hop sub-tasks, with the initial video-grounded hop acting as the primary bottleneck (\cf Table \ref{tab:different_qa}). More experimental findings are available in the supplementary material.

%% file: table_figs/tabBenchmarkCompare.tex
\begin{table*}[t]
\renewcommand{\arraystretch}{1.1}  
\centering
\caption{\textbf{Benchmark comparisons} across key dimensions: video domain scope, knowledge-driven focus, objective factuality focus, exclusivity of factual evaluation, multi-hop fact decomposition, and external evidence source.}
\scalebox{0.86}{
\begin{tabular}{@{}l c c c c c c@{}}
\toprule
\multirow{2}{*}{\textbf{Benchmarks}} & \textbf{Video} & \textbf{Knowledge}  & \textbf{Objective} & \textbf{Factuality} & \textbf{Multi-hop} & \textbf{Evidence} \\
 & \textbf{domain} & \textbf{driven} & \textbf{QA} & \textbf{exclusive} & \textbf{fact decomp.}  & \textbf{source} \\
\midrule
Video-MME \cite{fu2024video}  & Open & \xmark & \xmark & \xmark & \xmark  & \xmark \\
MMBench-Video \cite{fang2024mmbench}  & Open  & \xmark & \xmark & \xmark & \xmark  & \xmark  \\
Video-MMMU \cite{hu2025video}  & Professional & \cmark & \xmark & \xmark  & \xmark  & \xmark \\
MMVU \cite{zhao2025mmvu} & Discipline & \cmark & \xmark & \xmark & \xmark  & \cmark \\
MMWorld \cite{he2024mmworld} & Discipline & \cmark & \xmark & \xmark  & \xmark & \xmark \\
WorldQA \cite{zhang2024worldqa} & Open & \cmark & \xmark & \xmark & \xmark & \xmark\\
KnowIT-VQA \cite{garcia2020knowit} & TV shows & \cmark & \xmark & \xmark & \xmark & \xmark\\
\SimpleQA & Open & \cmark  & \cmark  & \cmark & \cmark  & \cmark \\
\bottomrule
\end{tabular}
}
\label{tab:benchComparison}
\end{table*}

%% file: sec/2_related.tex
\section{Related Work}\label{sec:2}
\input{table_figs/figDataPipeline}
\noindent \textbf{Factuality Benchmarks.} Factuality is the capability of LLMs to generate content that aligns with factual information, which can be substantiated by authoritative sources such as Wikipedia or textbooks \cite{akhtar2023multimodal,wang2024factuality}. Evaluating LLM factuality presents a non-trivial challenge and various benchmarks are proposed in the text-based \cite{lin2022truthfulqa,chern2023factool,gou2023critic,wei2024measuring,he2024chinese} and image-based scenarios \cite{marino2019ok,wang2017fvqa,jain2021select,gu2025see}. As one of the pioneering works, TruthfulQA \cite{lin2022truthfulqa} specifically targets imitative falsehoods in LLM-generated responses, which stem from erroneous preconceptions or knowledge gaps. Recently, the SimpleQA series of works \cite{wei2024measuring,he2024chinese,gu2025see,cheng2025simplevqa} facilitate factuality evaluation by constraining the scope to short, fact-seeking questions with simple answers, making factuality assessment more tractable compared to previous long, open-ended model outputs. Despite of this, the community urgently needs a standard benchmark for trustworthy factuality evaluation \emph{in video contexts}. To address this gap, our \SimpleQA emerges.

\noindent \textbf{Video Understanding Benchmarks.} Recently, video benchmarks have been designed for evaluations in comprehensive tasks, including temporal perception \cite{li2024mvbench}, reasoning \cite{cai2024temporalbench,chen2024autoeval}, navigation \cite{yang2024thinking}, long-form comprehension \cite{song2024moviechat,chandrasegaranhourvideo}, \etc. However, current video benchmarks largely overlook factuality evaluation, resulting in a lack of assessment for LVLMs' ability to generate factually accurate responses. Compared to video hallucination benchmarks \cite{wang2024videohallucer,guan2024hallusionbench,zhang2024eventhallusion}, which primarily assess models' adherence to video contents, our focused factuality evaluation emphasizes the model's alignment with verifiable external world knowledge \cite{wang2023survey,wang2024factuality}. 

\noindent \textbf{Differentiation from Knowledge-based and Discipline-base Benchmarks.}  Existing knowledge-based \cite{zhang2024worldqa,hu2025video} and discipline-based benchmarks \cite{zhao2025mmvu,he2024mmworld} either contain \emph{hypothetical/subjective} reasoning (\eg, the categories of \texttt{societal norms} and \texttt{social interactions} in WorldQA \cite{zhang2024worldqa}) or \emph{narrow their scopes} to single TV show \cite{garcia2020knowit} or discipline-related knowledge \cite{zhao2025mmvu,he2024mmworld}. Our \SimpleQA addresses these limitations by enforcing \emph{objective} factuality verification and ensuring \emph{diversity} across various categories. In addition, we introduce per-hop fact-grounded QA for \emph{fine-grained} evaluations.

%% file: table_figs/figDataPipeline.tex
\begin{figure*}[t]
	\centering
        \includegraphics[width=0.95\textwidth]{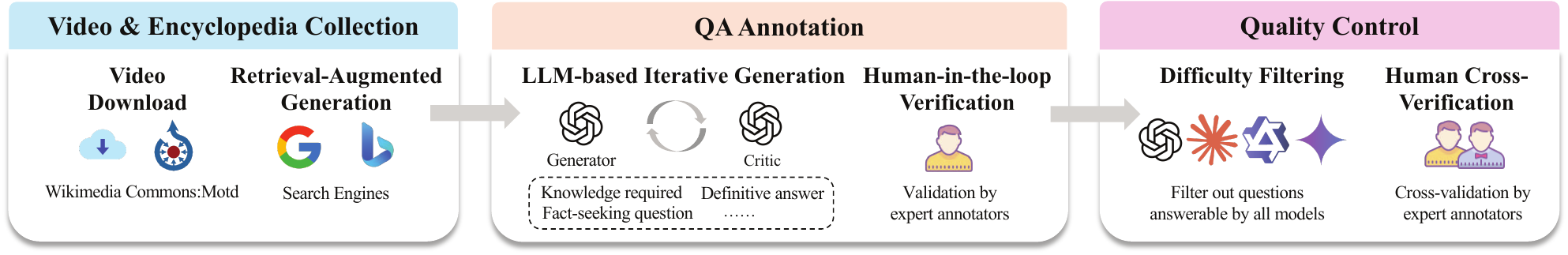}
	\caption{\textbf{An overview of the construction pipeline} of \SimpleQA.} 
	\label{fig:dataPipeline}
\end{figure*}

%% file: sec/3_method.tex
\section{Video SimpleQA} \label{sec:3}

\input{table_figs/figKnowledgeGen}

The construction pipeline of \SimpleQA is illustrated in Figure \ref{fig:dataPipeline}, which includes video \& encyclopedia collection, QA annotations, and quality control. 

\subsection{Video \& Encyclopedia Collection} 


\noindent \textbf{Video Collection}: To ensure broad coverage, we curate the knowledge-intensive videos from the ``Media of the Day" page of Wikimedia Commons\footnote{\url{https://commons.wikimedia.org/wiki/Commons:Media_of_the_day} \label{url}} together with the accompanied brief descriptions or scientific illustrations. Note that files on the ``Media of the Day" page are freely licensed, which avoids introducing any potential copyright concerns.

\noindent \textbf{Encyclopedia Collection}: As shown in Figure \ref{fig:knowledgeGen}, although the associated textual descriptions on the Wikimedia page provide related descriptions, the explanations for the specialized terms (\eg, \texttt{Barbary Ground Squirrel}, \texttt{Fuerteventura}) still lack formal definitions. To construct a more comprehensive encyclopedia, we leverage GPT-4o \cite{gpt4o} to extract key terms from the original descriptions and then obtain detailed explanations for these terms via RAG. Specifically, we apply LlamaIndex \cite{Liu_LlamaIndex_2022} as the RAG method, with search results from Google and Bing as data sources.

\subsection{QA Annotations} \label{sec:3.2}

The annotation pipeline for \SimpleQA follows a two-stage process: (1) automated LLM-based iterative generation and (2) human-in-the-loop verification refinement.

\noindent \textbf{LLM-based Iterative Generation:} The iterative generation process involves two LLMs, a \emph{generator} LLM for initial QA pair synthesis and a \emph{critic} LLM for quality assessment. Specifically, the generator receives videos and encyclopedic knowledge to generate candidate QA pairs. Subsequently, the critic LLM systematically evaluates output compliance with predefined quality criteria, providing targeted feedback for refinement. This iterative process continues for up to three refinement cycles, with non-compliant outputs being discarded post-final iteration to ensure rigorous quality control. Both generator and critic are implemented as GPT-4o. 

The explicit construction criteria are as follows: 1) \emph{Knowledge required}: The questions should necessitate both video content and relevant external factual knowledge. Those that can be answered solely based on either source should be excluded. For example, two questions that should be eliminated are: \texttt{What color is the insect in the video?} (which relies solely on video content) and \texttt{Which president of the United States was Obama?} (which relies solely on external knowledge); 2) \emph{Fact-seeking question}: The generated question should be factually grounded without any hypothetical or subjective reasoning; 3) \emph{Definitive answer}: To ensure a rigorous evaluation, each question must have a single, unambiguous, and time-invariant answer. To achieve this, we explicitly define the level of granularity in question phrasing. For example, we use ``which year" instead of ``when" and ``which city" instead of ``where" to eliminate ambiguity; 4) \emph{Short-form answer}: The answers should be in a short-form format; 5) \emph{Multi-hop facts}: To answer the question, it requires involving multiple factual sources; 6) \emph{Temporal grounded}: The questions are grounded in one or more video segments rather than specific frames.
\input{table_figs/tabMetrics}
\input{table_figs/tabStatistic}
\noindent \textbf{Human-in-the-loop Verification:} Through the iterative generation, we obtain QA annotations of reasonable quality. To further enhance the reliability, we train expert annotators to refine the LLM-generated QA annotations. The expert annotators are first required to watch the complete video and examine the collected encyclopedic knowledge. They then evaluate whether the LLM-generated QA annotations meet the specified criteria and manually revise them if necessary. 

To ensure \emph{multi-hop} fact grounding, annotators were additionally instructed to decompose each multi-hop QA into a series of sub-QAs (\cf Figure \ref{fig:vis}). The decomposition follows two rules: 1) \textbf{Single fact per sub-QA}: Each sub-QA targets a single fact that is independently verifiable; 2) \textbf{Referential dependency}: Each sub-QA builds upon the answer to the previous one, forming the step-by-step fact chaining.

\subsection{Quality Control} \label{sec:3.3}

\noindent \textbf{Difficulty Filtering.} To ensure an appropriate level of assessment difficulty, we establish filtering rules to exclude questions that are easy to answer. In particular, questions correctly answered by all four state-of-the-art models, including GPT-4o \cite{gpt4o}, Claude Sonnet 4 \cite{anthropic2025claude4}, Gemini 2.5 Pro \cite{comanici2025gemini}, and Qwen-VL-Max \cite{bai2023qwen} are deemed insufficiently challenging and consequently excluded from our benchmark. This filtering strategy ensures our dataset maintains a sufficient level of difficulty for meaningful evaluations.

\noindent \textbf{Human Cross-verification.} To further enhance the dataset quality, a rigorous human validation process is implemented. Each question is independently evaluated by two annotators for compliance with our predefined criteria. Questions are discarded if either annotator deems them non-compliant. Meanwhile, annotators are required to verify answers against authoritative sources (such as Wikipedia). Finally, the final dataset undergoes security auditing to address potential security issues. All these stringent human verification processes ensure both the accuracy of our dataset and its adherence to established criteria.



\noindent \textbf{Dataset Statistics.} The key statistics of \SimpleQA are demonstrated in Table \ref{tab:statistic}. As shown, it consists of 1079 videos with 1504 expert-annotated QA pairs. The videos span 4 primary categories, 15 secondary categories and 84 tertiary categories. The average lengths of questions and answers are 15.64 and 1.28 words, respectively, aligning with our intended short-form design. The video distribution at the secondary level is demonstrated in Figure \ref{fig:videoCategory}. The question type distribution is visualized in Figure \ref{fig:question_type}.

%% file: table_figs/figKnowledgeGen.tex
\begin{figure}[t]
	\centering
        \includegraphics[width=0.43\textwidth]{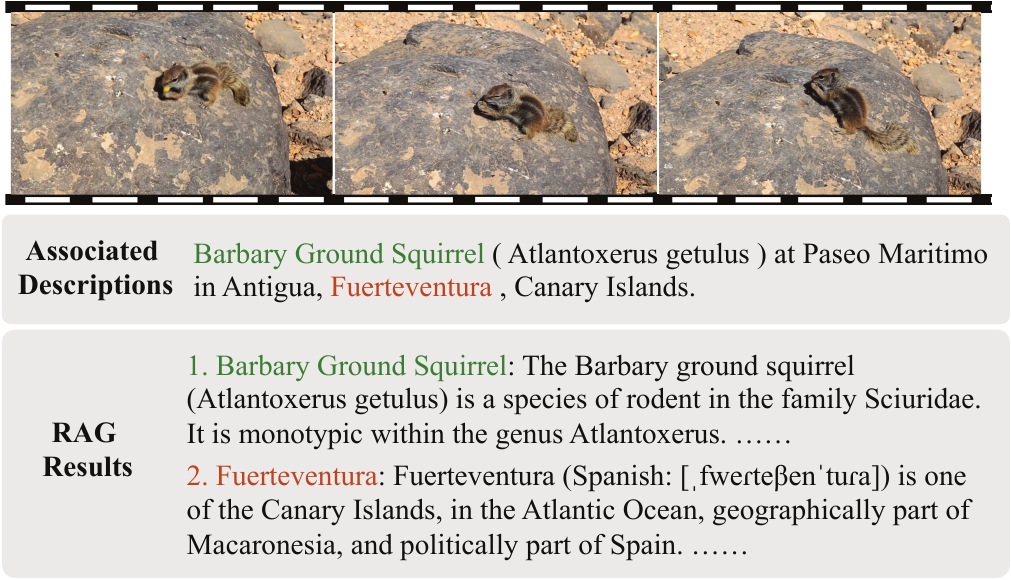}
	\caption{\textbf{The encyclopedia collection process} including the raw associated descriptions in Wikimedia and the RAG results\footref{url} for the specialized terms extracted by GPT-4o.}
	\label{fig:knowledgeGen}
\end{figure}

%% file: table_figs/tabMetrics.tex
\begin{table*}[!ht]
\centering
\setlength{\tabcolsep}{9pt}
\renewcommand\arraystretch{1.1}
\caption{\textbf{Evaluation results (\%) of open-source and proprietary multi-modal LLMs on \SimpleQA}. For metrics, CO, NA, IN, and CGA denote ``Correct", ``Not attempted", ``Incorrect", and ``Correct given attempted", respectively. For subtopics, ENG, NAT, SCI and SAC represent ``Engineering", ``Nature", ``Science" and ``Society and Culture".}
\scalebox{0.92}{
\begin{tabular}{lcccccccccc}
\toprule
\multirow{2}{*}{\textbf{Model}} & \multicolumn{5}{c}{\textbf{Overall results on 5 metrics}} & \multicolumn{4}{c}{\textbf{F-score on 4 primary categories}} \\
\cmidrule(r){2-6} \cmidrule(lr){7-10}
& CO & IN$\downarrow$ & NA$\downarrow$ & CGA & F-score & ENG & NAT & SCI & SAC  \\
\midrule
\multicolumn{10}{c}{\emph{\textbf{Human Performance}}}  \\ 
Human Open-book & 87.0 & 5.0 & 8.0 & 94.6 & 90.6
& 85.2 & 83.7 & 89.1 & 96.8 \\

Human Closed-book & 59.0 & 18.5 & 22.5 & 76.1 & 66.5
& 58.4 & 52.8 & 54.2 & 80.6 \\

\hline
\multicolumn{10}{c}{\emph{\textbf{Proprietary Multi-modal LLMs}}}  \\ 
o4-mini \cite{o3_o4mini} & 53.7 & 45.3 & 0.9 & 54.2 & 54.0 & 44.3 & 59.4 & 56.8 & 54.0 \\
o3 \cite{o3_o4mini} & \textbf{66.3} & \textbf{33.6} & \textbf{0.1} & \textbf{66.4} & \textbf{66.3} & \textbf{63.0} & \textbf{71.3} & 63.5 & \textbf{68.8} \\
GPT-4.5 \cite{gpt45} & 52.9 & 42.5 & 4.6 & 55.4 & 54.1 & 49.5 & 57.5 & 57.9 & 51.4 \\
GPT-4o \cite{gpt4o} & 47.7 & 45.9 & 6.4 & 51.0 & 49.3 & 45.1 & 57.1 & 52.7 & 45.4 \\
Claude Sonnet 4 \cite{anthropic2025claude4} & 32.8 & 51.2 & 16.0 & 39.0 & 35.6 & 33.0 & 34.3 & 37.9 & 35.0 \\
Claude 3.7 Sonnet \cite{anthropic2025claude3.7} & 32.6 & 47.3 & 20.1 & 40.8 & 36.2 & 24.2 & 40.3 & 41.9 & 34.5 \\
Claude 3.5 Sonnet2 \cite{anthropic2024claude3.5} & 33.7 & 58.2 & 8.1 & 36.7 & 35.2 & 26.1 & 37.2 & 38.5 & 35.4 \\
Claude 3.5 Sonnet \cite{anthropic2024claude3.5} & 31.5 & 53.5 & 15.0 & 37.0 & 34.0 & 26.8 & 36.0 & 38.1 & 32.5 \\
Gemini 2.5 Pro \cite{comanici2025gemini} & 61.2 & 34.3 & 4.5 & 64.1 & 62.6 & 53.5 & 65.8 & \textbf{67.1} & 61.5 \\
Gemini 2.5 Flash \cite{comanici2025gemini} & 53.7 & 34.9 & 11.3 & 60.6 & 57.0 & 46.0 & 61.6 & 61.4 & 56.2 \\
Qwen-VL-Max \cite{bai2023qwen} & 39.2 & 57.1 & 3.7 & 40.7 & 39.9 & 27.4 & 46.2 & 48.8 & 35.1 \\
Qwen-VL-Plus \cite{bai2023qwen} & 21.9 & 63.3 & 14.7 & 25.7 & 23.7 & 10.5 & 25.6 & 30.5 & 21.8 \\
\hline
\multicolumn{10}{c}{\emph{\textbf{Open-source Multi-modal LLMs}}}  \\
InternVL3-78B \cite{zhu2025internvl3} & 33.7 & 65.6 & \textbf{0.7} & 33.9 & 33.8 & 25.4 & 41.2 & 38.6 & 30.6 \\
InternVL3-38B \cite{zhu2025internvl3} & 31.4 & 67.7 & 0.9 & 31.7 & 31.5 & 21.3 & 33.3 & 35.7 & 31.8 \\
InternVL3-14B \cite{zhu2025internvl3} & 24.9 & 73.3 & 1.8 & 25.4 & 25.2 & 14.6 & 32.3 & 28.4 & 24.6 \\
InternVL3-9B \cite{zhu2025internvl3} & 22.6 & 72.9 & 4.5 & 23.7 & 23.1 & 12.8 & 33.2 & 27.9 & 19.9 \\
InternVL3-8B \cite{zhu2025internvl3} & 23.3 & 75.2 & 1.5 & 23.7 & 23.5 & 16.2 & 30.7 & 25.6 & 22.4 \\
\hdashline
Qwen2.5-VL-72B \cite{bai2025qwen2} & \textbf{38.7} & 57.3 & 4.0 & \textbf{40.3} & \textbf{39.5} & \textbf{26.1} & \textbf{47.0} & \textbf{48.2} & \textbf{34.7} \\
Qwen2.5-VL-32B \cite{bai2025qwen2} & 30.3 & 67.1 & 2.7 & 31.1 & 30.7 & 18.1 & 39.3 & 37.4 & 27.0 \\
Qwen2.5-VL-7B \cite{bai2025qwen2} & 24.7 & 71.2 & 4.1 & 25.8 & 25.3 & 13.8 & 25.6 & 30.8 & 25.1 \\
\hdashline
Qwen2-VL-72B \cite{wang2024qwen2} & 32.7 & 59.0 & 8.3 & 35.7 & 34.2 & 20.2 & 39.0 & 40.0 & 33.2 \\
Qwen2-VL-7B \cite{wang2024qwen2} & 22.4 & 69.4 & 8.2 & 24.4 & 23.4 & 15.9 & 23.9 & 25.1 & 25.0 \\
\hdashline
LLaVA-1.5-13B \cite{liu2023visual} & 19.3 & 76.7 & 4.1 & 20.1 & 19.7 & 11.6 & 21.2 & 21.9 & 20.8 \\
LLaVA-1.5-7B \cite{liu2023visual} & 16.1 & 78.5 & 5.4 & 17.1 & 16.6 & 8.9 & 19.2 & 19.0 & 17.1 \\
\hdashline
LLaVa-NeXT-Video-34B \cite{liu2024llavanext} & 11.2 & 83.8 & 4.9 & 11.8 & 11.5 & 7.6 & 11.5 & 10.3 & 14.5 \\
LLaVa-NeXT-Video-7B \cite{liu2024llavanext} & 9.3 & 52.9 & 37.8 & 14.9 & 11.4 & 7.4 & 15.1 & 14.5 & 8.7 \\
\hdashline
LLaVA-OneVision-72B \cite{li2024llava} & 25.4 & 73.6 & 1.0 & 25.7 & 25.5 & 15.9 & 25.3 & 28.5 & 27.3 \\
LLaVA-OneVision-7B \cite{li2024llava} & 18.9 & 76.6 & 4.5 & 19.8 & 19.3 & 12.1 & 26.3 & 21.3 & 18.4 \\
\hdashline
DeepSeek-VL2 \cite{wu2024deepseek} & 3.2 & 49.1 & 47.7 & 6.1 & 4.2 & 3.0 & 4.4 & 3.0 & 5.9 \\
DeepSeek-VL2-Small \cite{wu2024deepseek} & 5.9 & 52.1 & 42.1 & 10.1 & 7.4 & 3.9 & 10.1 & 9.5 & 6.2 \\
DeepSeek-VL2-Tiny \cite{wu2024deepseek} & 16.1 & 75.6 & 8.3 & 17.6 & 16.8 & 9.6 & 27.1 & 17.5 & 16.0 \\
\hdashline
Kimi-VL \cite{team2025kimi} & 18.3 & \textbf{44.4} & 37.3 & 29.1 & 22.4 & 14.9 & 19.8 & 25.2 & 24.5 \\
Keye-VL \cite{team2025kwai} & 25.4 & 53.9 & 20.7 & 32.0 & 28.3 & 15.3 & 23.1 & 37.2 & 26.6 \\

\bottomrule
\end{tabular}}
\label{tab:model-performance}
\end{table*}

%% file: table_figs/tabStatistic.tex
\begin{figure*}[t]
    \begin{subfigure}{0.30\textwidth}
        \centering
        \includegraphics[width=1.05\textwidth]{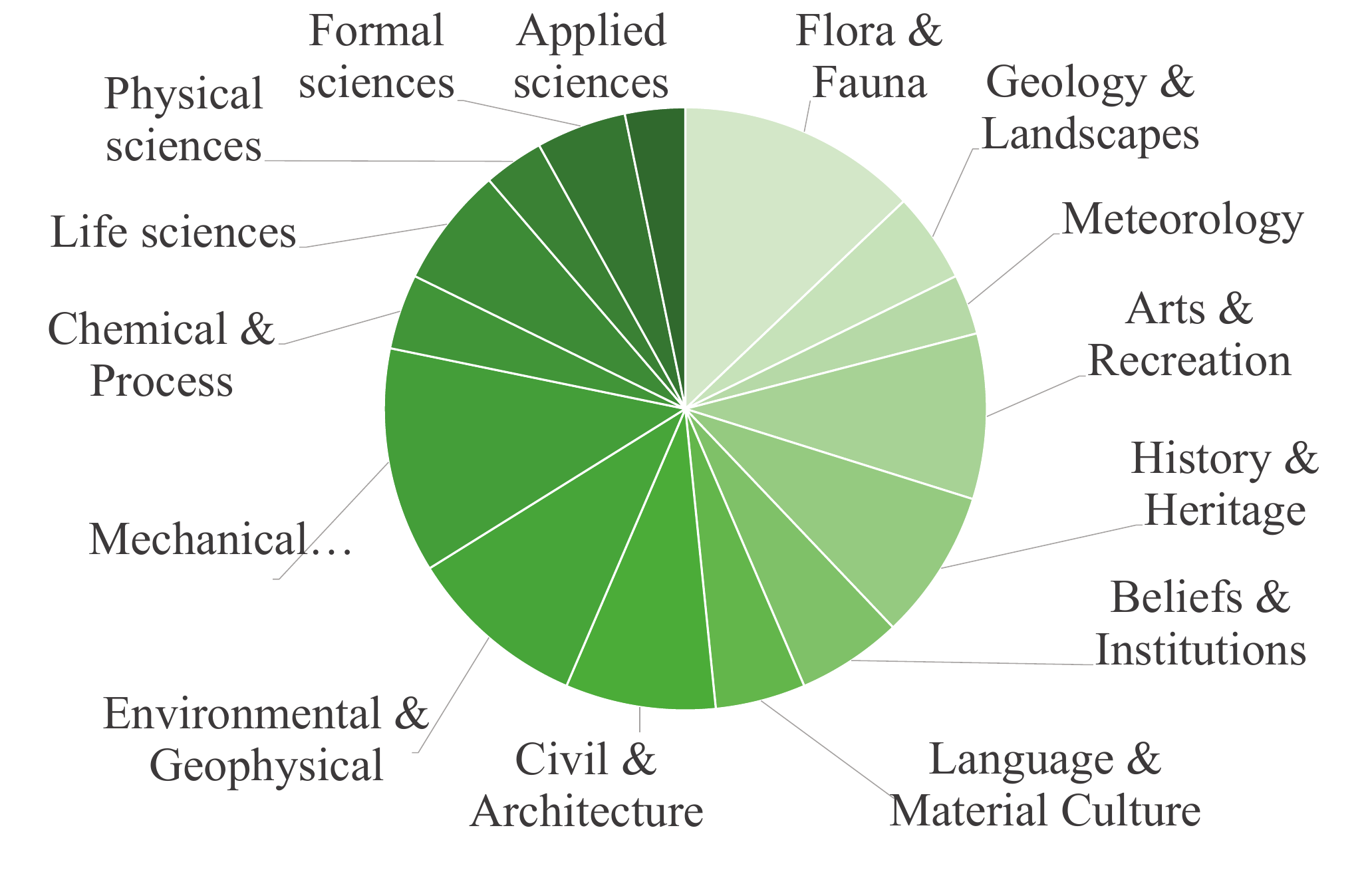} 
        \caption{}
        \label{fig:videoCategory}
    \end{subfigure}
    \hspace{-1mm}
    \begin{subfigure}{0.31\textwidth}
        \centering
        \includegraphics[width=1.05\textwidth]{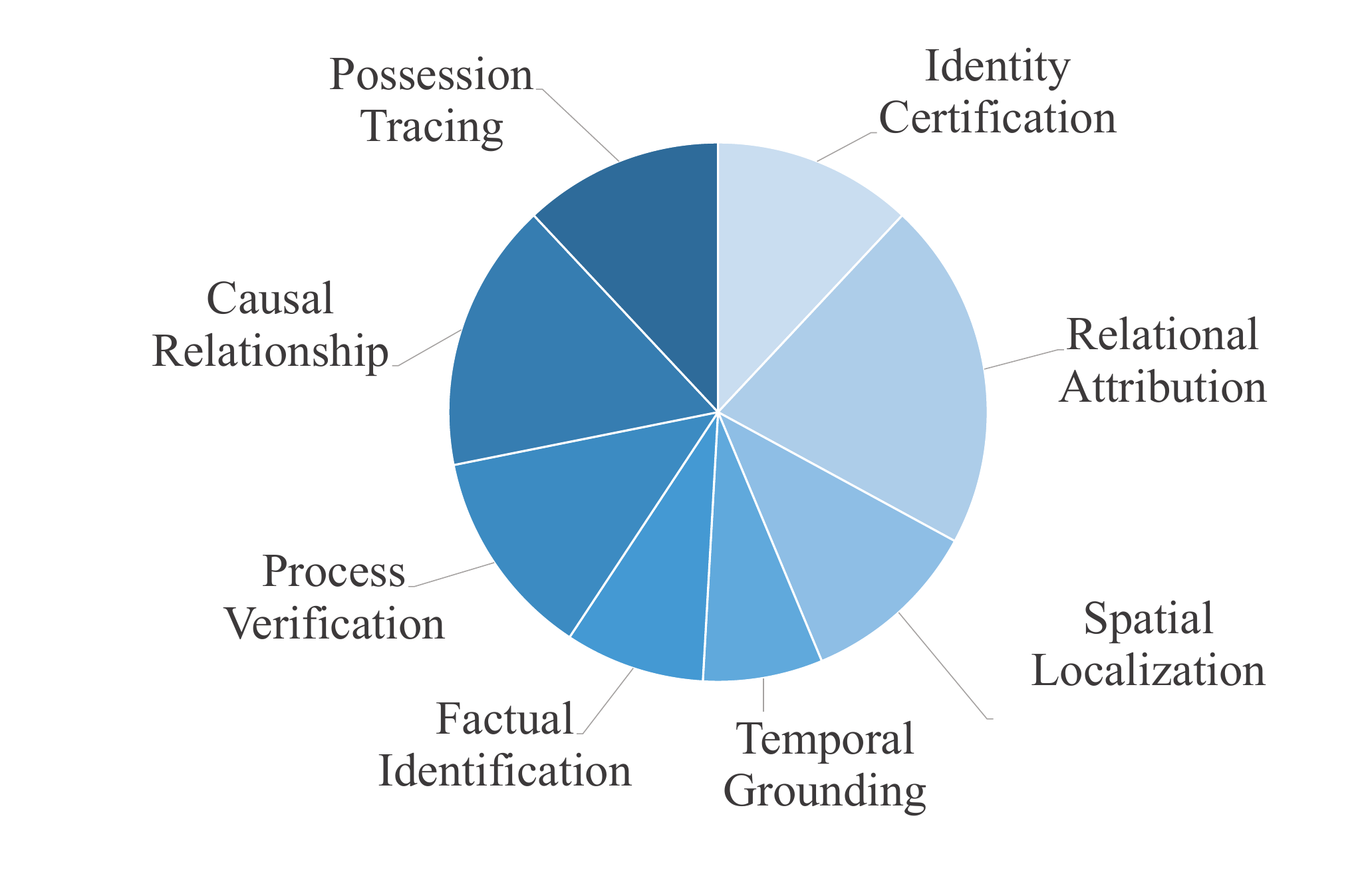} 
        \caption{}
        \label{fig:question_type}
    \end{subfigure}
    \hspace{1mm}
    \begin{subfigure}{0.3\textwidth}
        \centering
        \scalebox{0.83}{
        \begin{tabular}{lr}
        \toprule
        \textbf{Statistics of \SimpleQA} & \textbf{Value} \\
        \midrule
        2/3/4-hop QA pairs & 928 / 469 / 107 \\
        Total QA pairs & 1504 \\
        Question Length (avg/max) & 15.64 / 38 \\
        Answer Length (avg/max) & 1.28 / 9 \\
        Unique Videos & 1079 \\
        Video Length (Seconds, avg/max) & 201 / 8763 \\
        Number of primary category  & 4 \\
        Number of secondary category & 15 \\
        Number of tertiary category & 84 \\
        \bottomrule
        \end{tabular}}
        \caption{}
        \label{tab:statistic}
    \end{subfigure}
    \caption{(a) \textbf{Video distribution} at the secondary level; (b) \textbf{Question type distribution}; (c) \textbf{Key statistics}.}
    \label{fig:combined}
\end{figure*}

%% file: sec/4_experiments.tex
\section{Experiments} \label{sec:experiments}

\subsection{Experimental Setup}
\noindent \textbf{Evaluated Models.} We benchmark comprehensive state-of-the-art LLMs, including \textbf{12 proprietary models}, including o4-mini, o3, GPT-4.5, GPT-4o, Claude Sonnet 4, Claude 3.7 Sonnet, Claude 3.5 Sonnet series, Gemini 2.5 series, and Qwen-VL series, and \textbf{21 open-source models}, including InternVL3 series, Qwen2.5-VL series, Qwen2-VL series, LLaVA-1.5 series, LLaVA-NeXT-Video series, LLaVA-OneVision series, DeepSeek-VL2 series, Kimi-VL, and Keye-VL. Following Video-MME \cite{fu2024video}, we maximize frame utilization of each model by inputting the maximum frames that fit within its context window.

\noindent \textbf{Evaluation Metrics.} Following SimpleQA \cite{wei2024measuring}, we set up five evaluation metrics: (1) \textbf{Correct}: The predicted answer comprehensively contains all key information from the reference answer while containing no contradictory elements. (2) \textbf{Incorrect}: The predicted answer contradicts the reference answer. The indirect or equivocal responses (\eg, ``possibly", ``I think, although I'm not sure") are also considered incorrect. (3) \textbf{Not attempted}: The reference answer is not fully given in the predicted answer, and no statements in the answer contradict the gold target. (4) \textbf{Correct given attempted}: The ratio of correctly answered questions among attempted ones. (5) \textbf{F-score}: The harmonic mean values between \textit{correct} and \textit{correct given attempted} metrics. We follow the paradigm of \emph{LLM-as-a-Judge} \cite{gu2024survey} and employ \texttt{o3} as the judge model. 

\subsection{Experimental Findings\footnote{More experiments are available in the supplementary material.}}

The evaluation results on \SimpleQA are presented in Table \ref{tab:model-performance}, and key findings are summarized as follows:

\noindent \textbf{\SimpleQA is challenging:} To assess human performance on \SimpleQA, we sample 200 instances and recruit five participants to independently complete the tasks under two distinct conditions: with access to external resources (\eg, Internet, textbooks) and without such access. These configurations correspond to the \emph{human open-book} and \emph{human closed-book} settings documented in Table \ref{tab:model-performance}.

Compared to the human open-book performance, both open-source and proprietary models demonstrate suboptimal performance. Specifically, the top-performing proprietary model, o3 \cite{o3_o4mini}, achieves an F-score of 66.3\%. Open-source models exhibit even poorer results, with the best-performing one, Qwen2.5-VL-72B \cite{wang2024qwen2} attaining only 39.5\% F-score. This demonstrates that LVLMs still exhibit limited capability in factuality adherence within video contexts, while also highlighting the necessity of establishing \SimpleQA.

\input{table_figs/figErrorCaseHop4}
\input{table_figs/figCalibration}

\noindent \textbf{LVLMs are overconfident in what they generate:} All models exhibit higher IN values (incorrect predictions) than NA values (non-attempted responses), indicating a prevalent tendency to generate answers despite insufficient factual knowledge. To further investigate this overconfidence phenomenon, we conduct \emph{calibration} experiments \cite{guo2017calibration} to examine whether language models “know what they know”, \ie, whether the model’s assessed confidence scores align with the actual likelihood of its responses being correct. Specifically, we instruct LVLMs to self-assess confidence scores (0-100) for their predictions. Responses are grouped into confidence intervals (10-point bins), and we calculate \emph{interval accuracy} (correct predictions per bin). As shown in Figure \ref{fig:calibration}, except for o3, which demonstrates superior calibration, all other models \emph{mostly} fall below the perfect calibration line, indicating systematic overconfidence.
\input{table_figs/tabRAG}

\noindent \textbf{RAG yields significant gains at the cost of inference efficiency:} We explore RAG to facilitate \SimpleQA benchmark comprehension in a three-step approach: 1) Prompting GPT-4o with video and questions to extract key textual entities; 2) Applying LlamaIndex with Google and Wikipedia as sources to retrieve relevant information based on these extracted key entities; 3) Augmenting the input question with the retrieved information.

As shown in Table \ref{tab:rag}, RAG achieves consistent and significant F-score improvements over vanilla models. For instance, when integrated with Claude Sonnet 4 \cite{anthropic2025claude4}, RAG delivers an absolute improvement of 23.2\% (35.6\% \vs 58.8\%). However, this performance gain comes with substantial computational overhead. Table \ref{tab:rag} also quantifies the total inference time, demonstrating that RAG significantly impairs inference efficiency. Our findings highlight the critical trade-off between performance gains and computational practicality.
\input{table_figs/tabDifferentQA_4_hop}



\noindent \textbf{Per-hop factual evaluation:} In addition to multi-hop factual QA, our \SimpleQA benchmark also incorporates decomposed per-fact sub-QAs to facilitate fine-grained evaluations. As shown in Table 4, we present F-scores for both the final multi-hop QA and the sub-QAs across all 4-hop questions in \SimpleQA. Our analysis reveals: 1) \textbf{Multi-hop challenge}: The final multi-hop QAs achieve substantially lower F-scores than most single-hop sub-QAs, underscoring the difficulty of multi-hop fact grounding; 2) \textbf{First-hop bottleneck}: The F-score for the first hop is markedly lower than those of later hops, likely due to its reliance on accurate object or event recognition, which poses a key challenge. In contrast, LVLMs perform better on subsequent hops (Q2–Q4) given clearer contextual grounding.



In Figure \ref{fig:errorCaseHop4}, we visualize the per-hop evaluation results of the o3 model. This allows us to clearly identify which specific piece of factual knowledge the model lacks. For instance, in the left case, o3 lacks knowledge about the language family of English, while in the right case, it fails to recognize the film Caminandes: Llama Drama.

%% file: table_figs/figErrorCaseHop4.tex
\begin{figure*}[t]
	\centering
        \includegraphics[width=0.96\textwidth]{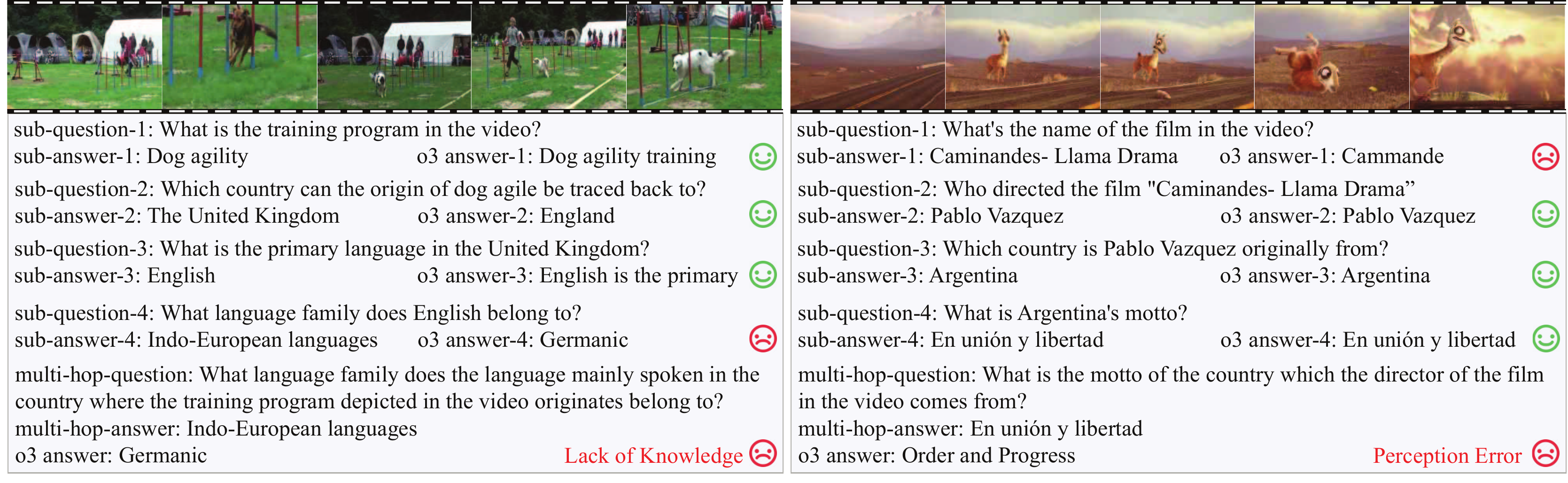}
	\caption{\textbf{Visualizations} of per-hop evaluation results of o3 \cite{o3_o4mini}.}
	\label{fig:errorCaseHop4}
\end{figure*}

%% file: table_figs/figCalibration.tex
\begin{figure}[t]
	\centering
        \includegraphics[width=0.43\textwidth]{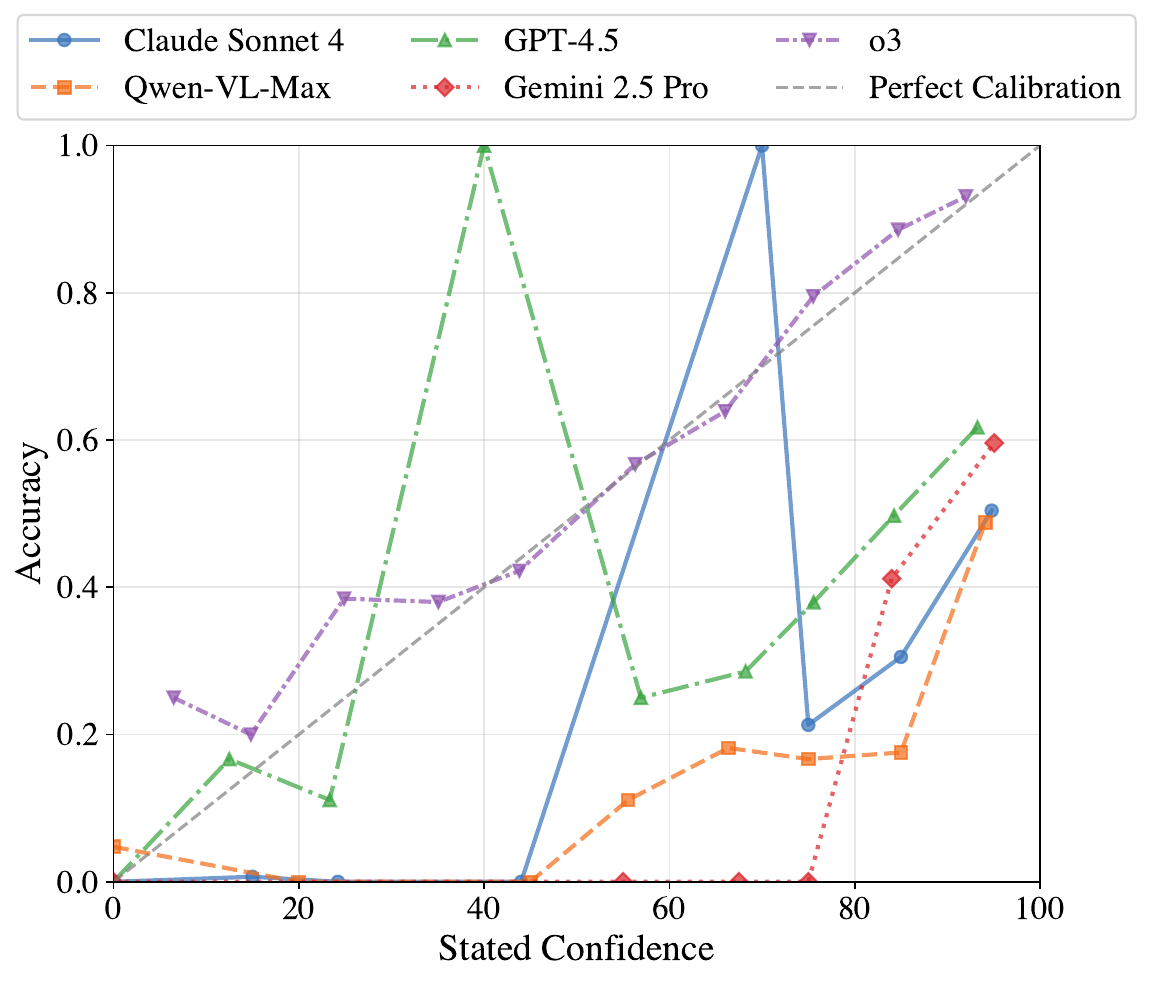}
	\caption{\textbf{Calibration curves} based on the self-stated confidence scores and interval-level accuracy.}
	\label{fig:calibration}
\end{figure}

%% file: table_figs/tabRAG.tex
\begin{table}[t]
\centering
\caption{\textbf{Comparisons between vanilla models and models with RAG} by F-score (\%) and the inference time (min).}
\scalebox{0.85}{
\begin{tabular}{lcccccccccc}
\toprule
\multirow{2}{*}{\textbf{Model}} & \multicolumn{2}{c}{\textbf{F-score}} & \multicolumn{2}{c}{\textbf{Inference Time}} \\
& vanilla & \emph{w/} RAG & vanilla & \emph{w/} RAG   \\
\midrule
o3 & 66.3 & \textbf{69.0} & \textbf{27.8} & 54.7 \\
GPT-4o & 49.3 & \textbf{61.3} & \textbf{30.1}  & 53.9\\
Gemini 2.5 Pro & 62.6 & \textbf{66.2} & \textbf{33.2} & 60.1\\
Claude Sonnet 4 & 35.6 & \textbf{58.8} & \textbf{29.9} & 56.4\\
Qwen-VL-Max & 39.9 & \textbf{57.0} & \textbf{24.2} & 61.2\\
\bottomrule
\end{tabular}}
\label{tab:rag}
\end{table}


%% file: table_figs/tabDifferentQA_4_hop.tex
\begin{table}[t]
\centering
\caption{\textbf{Per-hop factual evaluations} for 4-hop questions in terms of F-score (\%). Q1-Q4 denote the decomposed per-hop questions. Refer to Figure \ref{fig:errorCaseHop4} for an illustrative case.}
\scalebox{0.88}{
\begin{tabular}{lcccccccccc}
\toprule
\textbf{Model} & \textbf{QA1} &  \textbf{QA2} & \textbf{QA3} & \textbf{QA4} & \textbf{Multi-hop} \\
\midrule
o3 & \textbf{74.9} & \textbf{89.7} & \textbf{95.3} & \textbf{88.8} & \textbf{78.5} \\
GPT-4o & 59.1 & 85.4 & 92.5 & 80.9 & 47.0 \\
Claude Sonnet 4 & 43.5 & 85.7 & 80.8 & 61.6 & 47.7 \\
Gemini 2.5 Pro & 65.1 & 78.7 & 60.4 & 28.3 & 69.8 \\
Qwen-VL-Max & 37.9 & 69.2 & 83.6 & 72.3 & 40.7 \\
\bottomrule
\end{tabular}}
\label{tab:different_qa}
\end{table}

%% file: sec/5_con.tex
\section{Conclusions}
We present \texttt{Video SimpleQA}, the first benchmark explicitly designed for evaluating factual grounding in video contexts. Distinct from prior works, our framework introduces the following diagnostic dimensions: knowledge integration, multi-hop fact-seeking questioning, short-form definitive answering, and temporal grounded demands. Through the extensive evaluation of 33 state-of-the-art LVLMs, we reveal notable deficiencies in factual adherence, uncover prevalent model overconfidence, trade-offs associated with RAG, and the critical performance bottleneck.

%% file: sec/6_appendix.tex
\section{Supplementary Material} 
This \textbf{supplementary material} is organized as follows. We begin with a detailed description of the experimental setups, then present additional experimental findings, followed by the analysis of error types and the visualization results. 

Specifically, the \textbf{experiment setups} include the following aspects:
\begin{itemize}[topsep=2pt, partopsep=0pt, leftmargin=13pt, parsep=0pt, itemsep=3pt]
    \item Experimental configurations.
    \item Evaluation prompts.
    \item Video distributions. 
    \item Biographies of annotators.
    \item Copyright clarifications.
\end{itemize}
\textbf{More experimental findings} include the following aspects:
\begin{itemize}[topsep=2pt, partopsep=0pt, leftmargin=13pt, parsep=0pt, itemsep=3pt]
    \item Results with different judge models.
    \item Performance across secondary categories.
    \item Model size scaling remains effective.
    \item Frame number scaling remains effective.
    \item Performance of smaller LVLMs.
    \item Test-time compute yields limited benefits.
    \item Temporal scope analysis.
    \item More per-hop factual evaluation results.
\end{itemize}
\subsection{Experiment Setup} 

\noindent \textbf{Experimental configurations.} Table \ref{tab:model_configuration} details the configuration of each evaluated model. We used the default settings from the official implementation of each model. All experiments were reproducible on a workstation equipped with 8 NVIDIA A100 GPUs. Across all experiments, the temperature is set to 1.0, with a maximum output length of 1024 tokens.  Each experiment was repeated three times to ensure reproducibility and statistical reliability.

\noindent \textbf{Evaluation prompts.} The prompts for the grader, along with instructions guiding the model to output answers and confidence scores, are illustrated in Figure \ref{fig:grader_prompt2}, Figure \ref{fig:grader_prompt1} and Figure \ref{fig:answer_prompt}, respectively.

\noindent \textbf{Video distributions.} In Table \ref{tab:NA_EN_SC} and Table \ref{tab:SAC}, we present the video taxonomy distributions of \SimpleQA including 4 primary categories, 15 secondary categories, and 84 tertiary categories.

\noindent \textbf{Biographies of annotators.} The detailed information of annotators who participated in the construction of \SimpleQA can be found in Table \ref{tab:annotators}. All annotators come from universities ranked in the top 500 of the 2026 QS Global Rankings\footnote{\url{https://www.topuniversities.com/world-university-rankings} \label{url_ranking}}, and they are all fluent in English.

\noindent \textbf{Copyright clarifications.} The videos of our \SimpleQA benchmark are sourced from the ``Media of the Day" page of Wikipedia, which are freely licensed. Therefore, the construction of \SimpleQA \emph{avoids introducing any potential copyright concerns}.

To better align with ethical standards, we have implemented the following measures: 1) \textbf{Data desensitization}: To reduce possible security threats during the evaluation stage, the finalized dataset was independently reviewed by six professional security auditors. Each data instance was examined by at least two reviewers, focusing on potential issues such as gender bias, offensive language, or politically sensitive content. Only the samples that passed all checks were included in the final release; 2) \textbf{Non-commercial research}: We release our code and dataset under a Creative Commons Attribution-NonCommercial-ShareAlike 4.0 International (CC BY-NC-SA 4.0) License. They are available strictly for non-commercial research; 3) \textbf{Opt-out mechanism}: We release our dataset with an opt-out mechanism, allowing content owners and individuals appearing in the videos to request the removal of their video references.

\subsection{More Experimental Findings} 
\input{table_figs/figRobustness}
\input{table_figs/tabEvalModels_Configuration}
\input{table_figs/figFscoreOnSubtopic}
\input{table_figs/figModelSize}

\noindent \textbf{Results with different judge models.} The short-form answer paradigm of \SimpleQA enables automated evaluation through LLM-as-a-judge frameworks with low run-to-run variance. To demonstrate this, we select five typical LVLMs, and evaluate them using various judge models including o3 \cite{o3_o4mini}, Gemini 2.5 Pro \cite{comanici2025gemini}, Gemini 2.5 Flash \cite{comanici2025gemini}, Qwen-VL-Max \cite{bai2023qwen}, and Claude Sonnet 4 \cite{anthropic2025claude4}. As shown in Figure \ref{fig:robustness}, while the specific scores from different judge models vary, the relative rankings of the evaluated models remain consistent. This demonstrates the robustness of our evaluation under the short-form paradigm.

\noindent \textbf{Performance across secondary categories.} Figure \ref{fig:fscore_on_subtopic} demonstrates the F-score performance across 15 secondary categories. As shown, we observe distinct performance patterns among the compared LVLMs. 
\begin{itemize}[topsep=2pt, partopsep=0pt, leftmargin=13pt, parsep=0pt, itemsep=3pt]
    \item \emph{Capability distribution}: o3 \cite{o3_o4mini} demonstrates the most consistent performance with superior F-scores across domains, particularly excelling in the category of \texttt{Meteorology}. Gemini 2.5 Pro \cite{comanici2025gemini} follows with complementary strengths and exhibits notable advantages in the category of \texttt{Life sciences} compared to other LVLMs.

     \item \emph{Imbalanced performance profiles}: Qwen2.5-VL-72B \cite{bai2025qwen2} and Qwen-VL-MAX \cite{bai2023qwen} show significant performance variance, with Qwen-VL-MAX \cite{bai2023qwen} severely underperforming in \texttt{Formal Sciences} compared to its moderate capabilities in other domains. Kimi-VL \cite{team2025kimi} and Keye-VL \cite{team2025kwai} consistently underperform across most domains, with particularly low F-scores in the technical areas (\eg, \texttt{Physical} \texttt{Sciences}, \texttt{Applied} \texttt{Sciences}).

   \item \emph{Disciplinary performance gap}: All models exhibit systematically lower F-scores in scientific domains (\eg, \texttt{Formal/Applied Sciences}) compared to humanities-oriented categories (History \& Heritage, Civil \& Architecture), with Kimi-VL \cite{team2025kimi} showing the most pronounced disparities.
\end{itemize}

\noindent \textbf{Model size scaling remains effective:} As evidenced by the F-score across various model sizes in Figure \ref{fig:modelSize}, model size scaling continues to demonstrate effectiveness, where larger architectures exhibit consistently superior performance. This observation aligns with the widely recognized scaling law principle.

\noindent \textbf{Frame number scaling remains effective:} In Figure \ref{fig:ablateFrame}, we demonstrate the impact of the number of sampled frames on performance. The results for Gemini 2.5 Pro, o3, and Gemini 2.5 Flash reveal a positive correlation between the video frame number and the F-score, thereby validating the effectiveness of frame number scaling. For the results of Claude Sonnet 4, the optimum result is achieved when setting the frame number to 10.
\input{table_figs/figAblateFrame}

\noindent \textbf{Performance of smaller LVLMs:} In addition to the evaluation results of the 33 LVLMs presented in the main paper, we have also included results for smaller models (with parameter counts of 0.5B, 1B, 2B and 3B) in Table \ref{tab:model-performance-small}. This serves as a valuable reference for users with limited resources who wish to utilize the proposed \SimpleQA dataset.
\input{table_figs/figTTC}

\noindent \textbf{Test-Time compute yields limited benefits}: We empirically investigate the effectiveness of test-time compute \cite{snell2024scaling} strategies on \SimpleQA by conducting experiments with 200 randomly sampled instances. Two approaches are evaluated: 1) \textbf{Best-of-N}: The model independently generates $N$ responses and selects the one it considers the best as the final answer; 2) \textbf{Self-refine}: The model is prompted to iteratively refine the initial outputs using self-generated feedback \cite{madaan2023self}. 

Figure \ref{fig:TTC} presents the accuracy (\ie, the proportion of correct answers) under varying inference-time $N$ (for Best-of-N) and different refinement iterations (for Self-refine). Experimental results reveal that both strategies fail to produce significant or consistent accuracy improvements. In some cases, these strategies even degrade performance. For instance, when increasing $N$ from 2 to 4 in Best-of-N trials, o3 \cite{o3_o4mini} and Gemini-2.5-Pro \cite{reid2024gemini} exhibit reduced accuracy, suggesting that these models struggle to reliably select the best answer from multiple inferences. These findings highlight the challenges in improving factuality through post-hoc test-time compute strategies.
\input{table_figs/tabMetricsSmall}
\input{table_figs/tabTemporal}

\noindent \textbf{Temporal scope analysis:} \SimpleQA is designed to be temporally grounded, \ie, answering questions should refer to one or more temporal segments in the video, rather than relying on a single frame. To demonstrate this, we randomly sampled 200 QA pairs and instructed expert annotators to categorize them based on the \emph{necessary temporal scope} needed for accurate answers. We categorize the results as follows: 1) \emph{short-term} scope required ($<$10 seconds); 2) \emph{medium-term} scope required (10s-1min); or 3) \emph{long-term} scope required ($>$1min). It should be emphasized that our definitions of short/medium/long-term specifically denote the temporal scope required to correctly answer QA pairs, distinct from the absolute video duration referenced in existing long-form video understanding benchmarks \cite{wang2024lvbench,fu2024video}. 

Table \ref{tab:temporal} summarizes the proportion and performance under different temporal scopes, which reveals that 82.6\% of cases require either short-term or medium-term temporal understanding to answer correctly. This distribution demonstrates that our \SimpleQA benchmark indeed necessitates temporal understanding capabilities rather than simple frame-level analysis. Furthermore, as indicated in Table \ref{tab:temporal}, videos with long-term temporal scope exhibit significantly lower performance metrics compared to the other two categories, which highlights the importance of long-context temporal modeling.

\noindent \textbf{More per-hop factual evaluation results:} In the main paper, we have presented the performance of per-fact sub-QA for all the 4-hop questions \SimpleQA. Here, we present the per-fact sub-QA performance across all 3-hop questions. As shown in Table \ref{tab:different_qa_new}, we still observe that final multi-hop QAs achieve substantially lower F-scores than most single-hop sub-QAs, and the performance of the first hop is markedly lower than that of subsequent hops. 

We observe that the accuracy of multi-hop questions is higher than that of sub-QA1. Through analysis, we found this occurs because in certain cases, even if sub-QA1 is answered incorrectly, both the ground-truth answer and the incorrect answer lead to a consistent final answer for the multi-hop question. For example in Figure \ref{fig:randomguess}, for the model o3, the \texttt{kata} type is incorrectly identified as \texttt{Heian Nidan}. Yet, whether based on this incorrect answer or the correct answer (\texttt{Heian Godan}), the final answer for the multi-hop question is consistently \texttt{Japan}. This further highlights the value of our proposed fine-grained per-hop evaluation, \ie, it requires the model not only to correctly answer the final multi-hop question, but also to accurately answer each single-hop, fact-based question.
\input{table_figs/figDistributionError}
\subsection{Error Analysis} 

\input{table_figs/figRandomGuessCase}

\input{table_figs/figErrorCase}
This section presents a case study analyzing error patterns in o3 \cite{o3_o4mini}, Gemini 2.5 Pro \cite{comanici2025gemini} and Qwen2.5-VL-72B \cite{bai2025qwen2} through a systematic examination of all available samples per model spanning diverse question types. We categorize observed errors into four primary classes, each illustrated with representative examples:
\begin{itemize}[topsep=2pt, partopsep=0pt, leftmargin=13pt, parsep=0pt, itemsep=3pt]
\item  \textbf{Perception error:} Incorrect identification of objects. This occurs when LVLMs misidentify or fail to detect key visual elements in input videos (\cf Figure \ref{fig:errorCase1} (a)).

\item \textbf{Lack of knowledge:} Correct identification but lacking relevant knowledge. LVLMs accurately perceive the visual content but cannot provide accurate information due to knowledge limitations (\cf Figure \ref{fig:errorCase1} (b)).

\item \textbf{Refusal to answer:} LVLMs recognize their inability to make a confident determination and opt to abstain from answering (\cf Figure \ref{fig:errorCase2} (c)).

\item \textbf{Failure to follow instructions:} LVLMs understand the input but fail to properly execute the given instructions. This typically manifests in two ways: 1) the generated outputs do not conform to the specified format requirements, or 2) the responses are irrelevant to the question posed, \eg, addressing different topics than what was actually requested (\cf Figure \ref{fig:errorCase2} (d)).
\end{itemize}

\input{table_figs/tabDifferentQA_3_hop}


\subsection{Visualizations} 

\noindent \textbf{Evaluation results}: The visualization results of three typical LVLMs (o3 \cite{o3_o4mini}, Gemini 2.5 Pro \cite{comanici2025gemini} and Qwen2.5-VL-72B \cite{bai2025qwen2}) are illustrated in Figure \ref{fig:viscase1} – Figure \ref{fig:viscase5}.

\noindent \textbf{Limitations of existing discipline-based methods}:  As analyzed in the main paper, current discipline-based methods suffer from the following issues: 1) \emph{Inclusion of hypothetical or subjective speculation}: For instance, in Figure \ref{fig:subjectiveCase}, the questions often contain hypothetical or subjective expressions such as ``might", ``infer", ``seem to", or ``probably". In contrast, our \SimpleQA requires answers to be crafted in an unambiguous and definitively correct manner; 2) \emph{Entangling fact grounding and reasoning}: To answer the questions in the examples shown in Figure \ref{fig:reasoningCase}, the model needs to possess reasoning abilities such as matrix computation, minimum spanning tree planning, and time complexity analysis. This makes it difficult to independently evaluate the model’s fact-grounding capability. In contrast, our \SimpleQA exclusively focuses on fact identification, providing a clearer assessment of LVLMs’ fact-grounding ability.


\input{table_figs/figGraderPrompt}
\input{table_figs/figAnswerPrompt}
\input{table_figs/tabAnnotatorList}
\input{table_figs/tabCountOfNA_EN_SC}
\input{table_figs/tabCountOfSAC}

\input{table_figs/subjectiveCase}

\input{table_figs/reasoningCase}


\input{table_figs/figVisCase}

%% file: table_figs/figRobustness.tex
\begin{figure}[t]
	\centering
 \includegraphics[width=0.5\textwidth]{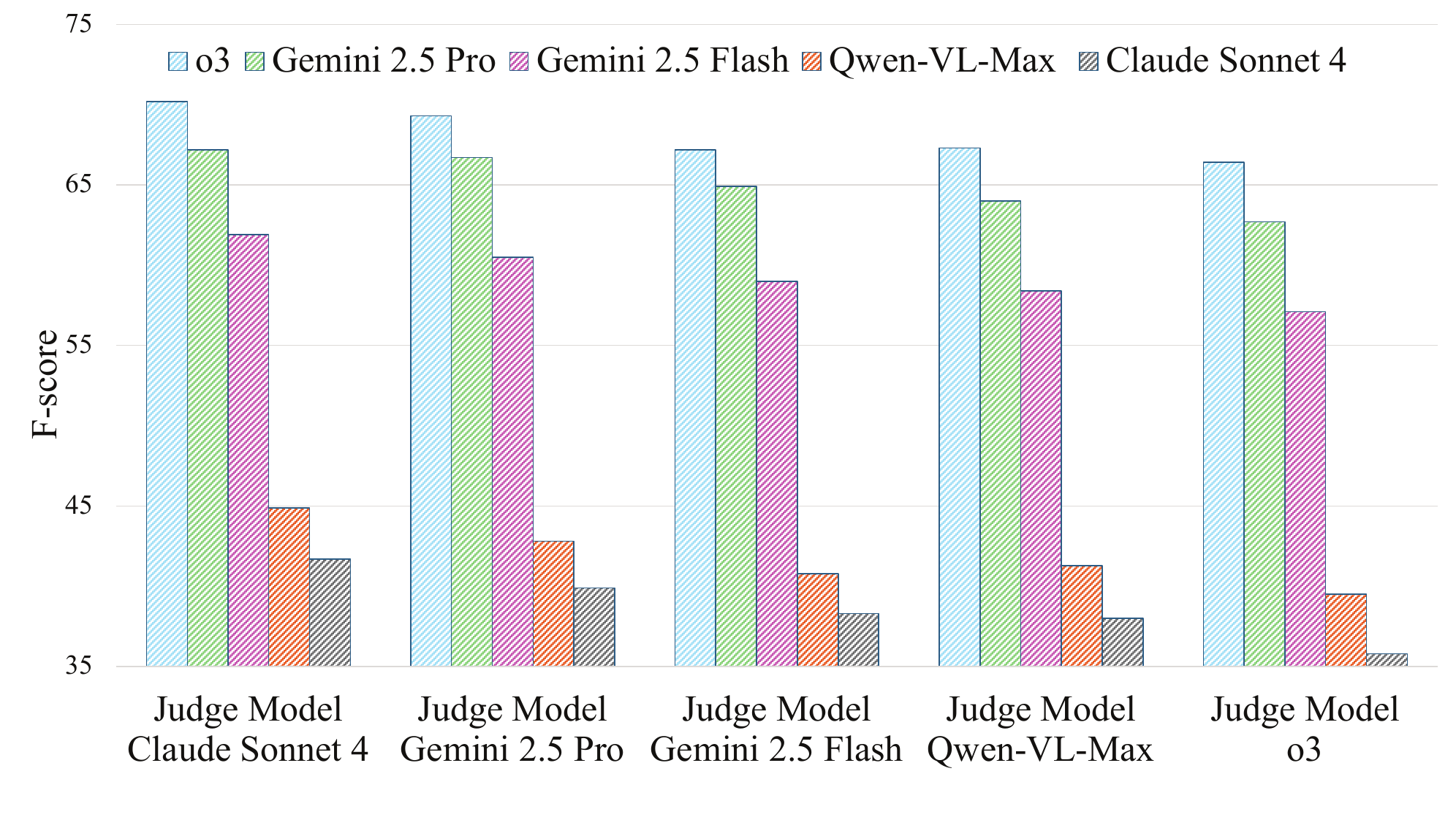}
	\caption{Results with different judge models.}
	\label{fig:robustness}
\end{figure}

%% file: table_figs/tabEvalModels_Configuration.tex
\begin{table*}[!ht]
\centering
\footnotesize
\caption{\textbf{Configurations} of evaluated LVLMs in \SimpleQA.}
\vspace{-1mm}
\scalebox{0.99}{
\begin{tabular}{lllcrc}
\toprule
\textbf{Organization} & \textbf{Model} & \textbf{Release} & \textbf{Version} & \textbf{\begin{tabular}[c]{@{}c@{}}Input\\Frames\end{tabular}} & \\
\midrule
\multicolumn{6}{c}{\emph{\textbf{Proprietary Multi-modal LLMs}}} \\
 \midrule
\multirow{4}{*}{OpenAI} 
& o4-mini & 2025-4 & \texttt{o4-mini-2025-04-16} & 32\\
& o3 & 2025-4 & \texttt{o3-2025-04-16} & 32\\
& GPT-4.5 &  2025-2  &  \texttt{gpt-4.5-2025-02-27} & 32  \\
& GPT-4o &  2024-5  &  \texttt{gpt-4o-2024-08-06} & 32  \\

\noalign{\vskip 0.5ex}\hdashline\noalign{\vskip 0.5ex}
\multirow{4}{*}{Anthropic}
& Claude Sonnet 4 & 2025-5 & \texttt{claude-sonnet-4} & 32 \\
& Claude 3.7 Sonnet & 2025-2 & \texttt{claude-3.7-sonnet} & 16 \\
& Claude 3.5 Sonnet2 & 2024-10 & \texttt{claude-3.5-sonnet-v2} & 16 \\
& Claude 3.5 Sonnet & 2024-6 & \texttt{claude-3.5-sonnet} & 16 \\

\noalign{\vskip 0.5ex}\hdashline\noalign{\vskip 0.5ex}

\multirow{2}{*}{Google} 
& Gemini 2.5 Pro & 2025-3 & \texttt{gemini-2.5-pro-2025-06-17} & 32 & \\
& Gemini 2.5 Flash & 2025-3 & \texttt{gemini-2.5-flash-2025-06-17} & 32\\

\noalign{\vskip 0.5ex}\hdashline\noalign{\vskip 0.5ex}

\multirow{2}{*}{Alibaba} 
& Qwen-VL-Max & 2024-1 & \texttt{qwen-vl-max} & 32\\
& Qwen-VL-Plus & 2023-11 & \texttt{qwen-vl-plus} & 32\\

\midrule
\multicolumn{6}{c}{\emph{\textbf{Open-source Multi-modal LLMs}}} \\
\midrule

\multirow{7}{*}{Shanghai AI Lab}
& InternVL3-78B & 2025-4 & \texttt{InternVL3-78B} & 16 \\
& InternVL3-38B & 2025-4 & \texttt{InternVL3-38B} & 16 \\
& InternVL3-14B & 2025-4 & \texttt{InternVL3-14B} & 16 \\
& InternVL3-9B & 2025-4 & \texttt{InternVL3-9B} & 16 \\
& InternVL3-8B & 2025-4 & \texttt{InternVL3-8B} & 16 \\
& InternVL3-2B & 2025-4 & \texttt{InternVL3-2B} & 16 \\
& InternVL3-1B & 2025-4 & \texttt{InternVL3-1B} & 16 \\

\noalign{\vskip 0.5ex}\hdashline\noalign{\vskip 0.5ex}

\multirow{7}{*}{Alibaba} 
& Qwen2.5-VL-72B & 2025-2 & \texttt{Qwen2.5-VL-72B-Instruct} & 32 \\
& Qwen2.5-VL-32B & 2025-2 & \texttt{Qwen2.5-VL-32B-Instruct} & 32 \\
& Qwen2.5-VL-7B & 2025-2 & \texttt{Qwen2.5-VL-7B-Instruct} & 32 \\
& Qwen2.5-VL-3B & 2025-2 & \texttt{Qwen2.5-VL-3B-Instruct} & 32 \\
& Qwen2-VL-72B & 2024-8 & \texttt{Qwen2-VL-72B-Instruct} & 16 \\
& Qwen2-VL-7B & 2024-8 & \texttt{Qwen2-VL-7B-Instruct} & 16 \\
& Qwen2-VL-2B & 2024-8 & \texttt{Qwen2-VL-2B-Instruct} & 16 \\

\noalign{\vskip 0.5ex}\hdashline\noalign{\vskip 0.5ex}

\multirow{7}{*}{Llava Hugging Face} 
& LLaVA-1.5-13B & 2023-9 & \texttt{llava-1.5-13b-hf} & 4 \\
& LLaVA-1.5-7B & 2023-9 & \texttt{llava-1.5-7b-hf} & 4 \\
& LLaVA-NeXT-Video-34B & 2024-4 & \texttt{LLaVA-NeXT-Video-34B-hf} & 8 \\
& LLaVA-NeXT-Video-7B & 2024-4 & \texttt{LLaVA-NeXT-Video-7B-hf} & 4 \\
& LLaVA-OneVision-72B & 2024-8 & \texttt{llava-onevision-qwen2-72b-ov-hf} & 16 \\
& LLaVA-OneVision-7B & 2024-8 & \texttt{llava-onevision-qwen2-7b-ov-hf} & 16 \\
& LLaVA-OneVision-0.5B & 2024-8 & \texttt{llava-onevision-qwen2-0.5b-ov-hf} & 16 \\

\noalign{\vskip 0.5ex}\hdashline\noalign{\vskip 0.5ex}

\multirow{3}{*}{DeepSeek} 
& DeepSeek-VL2 & 2024-12 & \texttt{deepseek-vl2} & 2 \\
& DeepSeek-VL2-Small & 2024-12 & \texttt{deepseek-vl2-small} & 2\\
& DeepSeek-VL2-Tiny & 2024-12 & \texttt{deepseek-vl2-tiny} & 2\\

\noalign{\vskip 0.5ex}\hdashline\noalign{\vskip 0.5ex}

\multirow{1}{*}{MoonshotAI} 
& Kimi-VL & 2025-4 & \texttt{Kimi-VL-A3B-Instruct} & 32 \\

\noalign{\vskip 0.5ex}\hdashline\noalign{\vskip 0.5ex}

\multirow{1}{*}{Kwai-Keye} 
& Keye-VL & 2025-6 & \texttt{Keye-VL-8B-Preview} & 32 \\

\bottomrule
\end{tabular}
}
\label{tab:model_configuration}
\end{table*}

%% file: table_figs/figFscoreOnSubtopic.tex
\begin{figure}[t]
	\centering
        \includegraphics[width=0.5\textwidth]{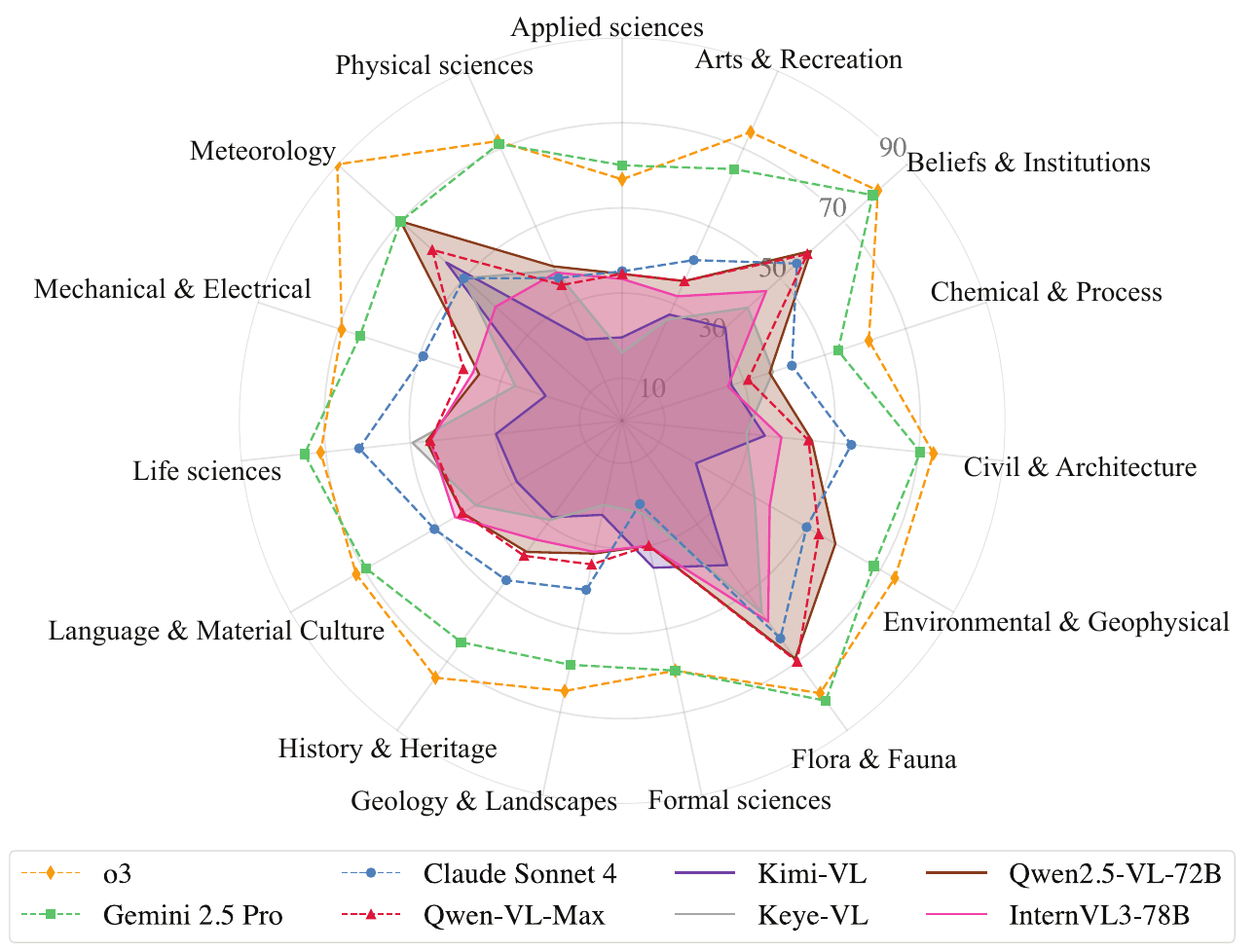}
	\caption{The performance of different models across 15 secondary categories in \SimpleQA.}
	\label{fig:fscore_on_subtopic}
\end{figure}

%% file: table_figs/figModelSize.tex
\begin{figure}[t]
	\centering
        \includegraphics[width=0.45\textwidth]{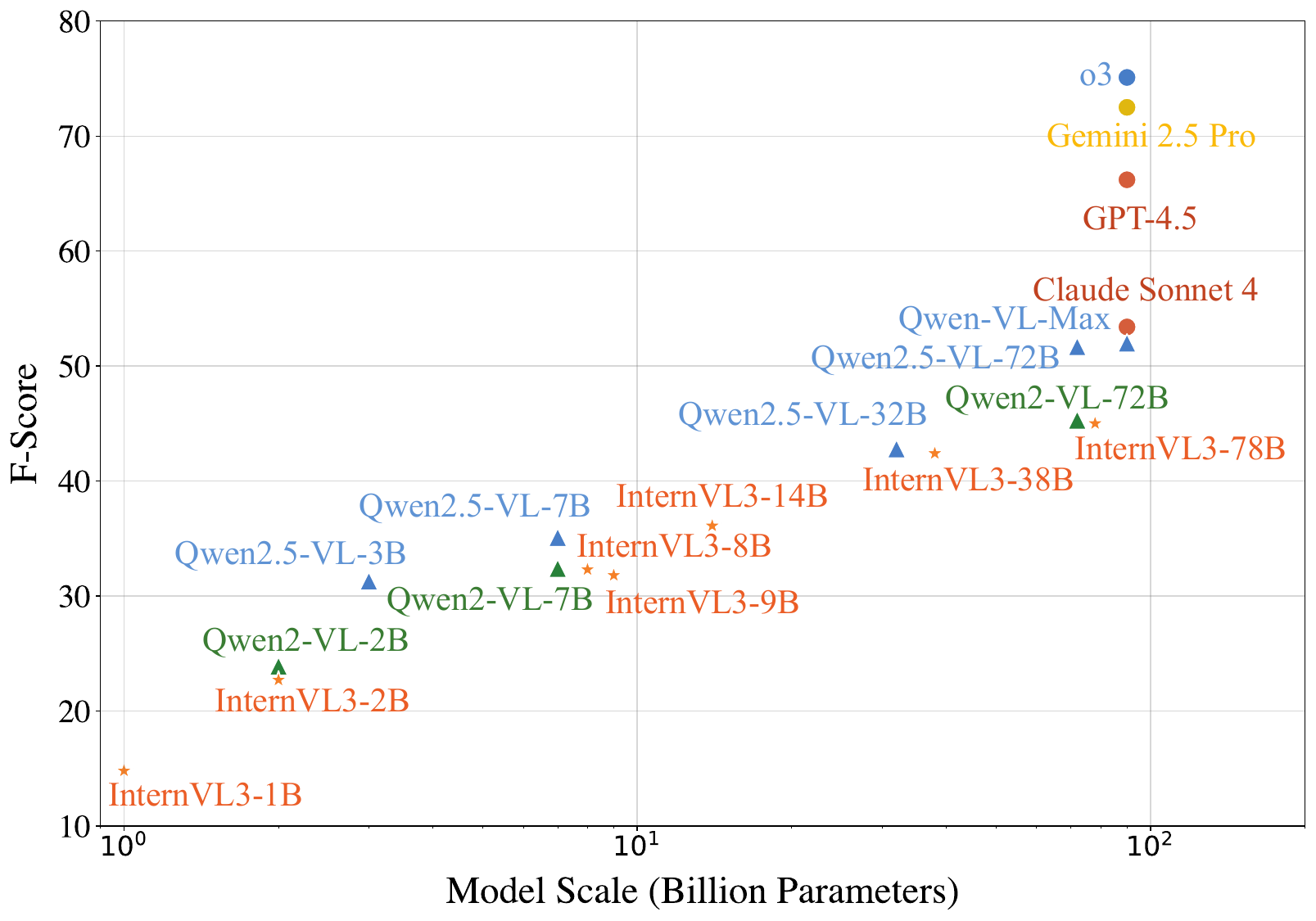}
	\caption{The relationship between model size and F-score.}
	\label{fig:modelSize}
\end{figure}

%% file: table_figs/figAblateFrame.tex
\begin{figure}[t]
	\centering
        \includegraphics[width=0.48\textwidth]{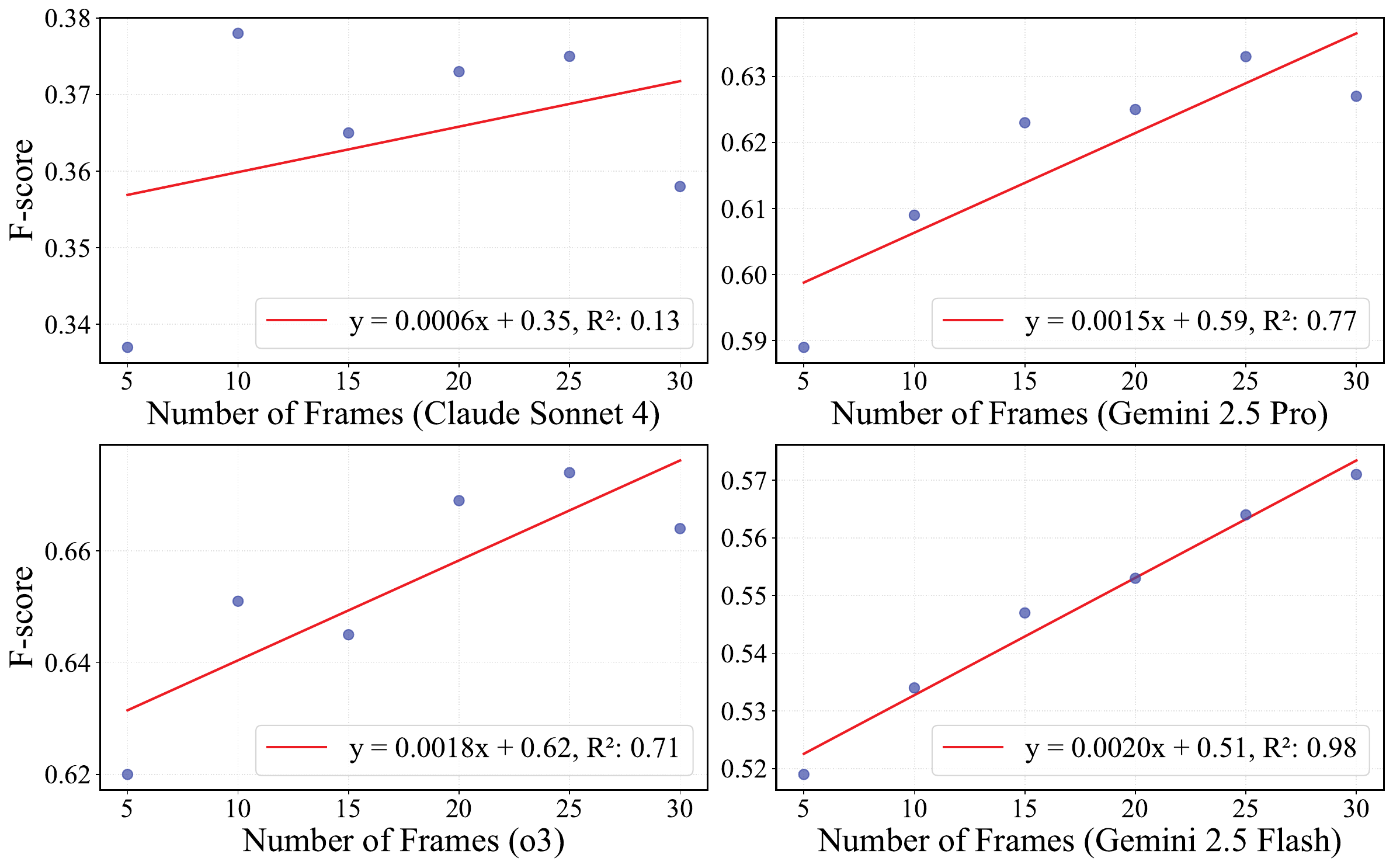}
	\caption{Frame number scaling experiments.}
	\label{fig:ablateFrame}
\end{figure}

%% file: table_figs/figTTC.tex
\begin{figure}[t]
	\centering
        \hspace{-1mm}
        \includegraphics[width=0.49\textwidth]{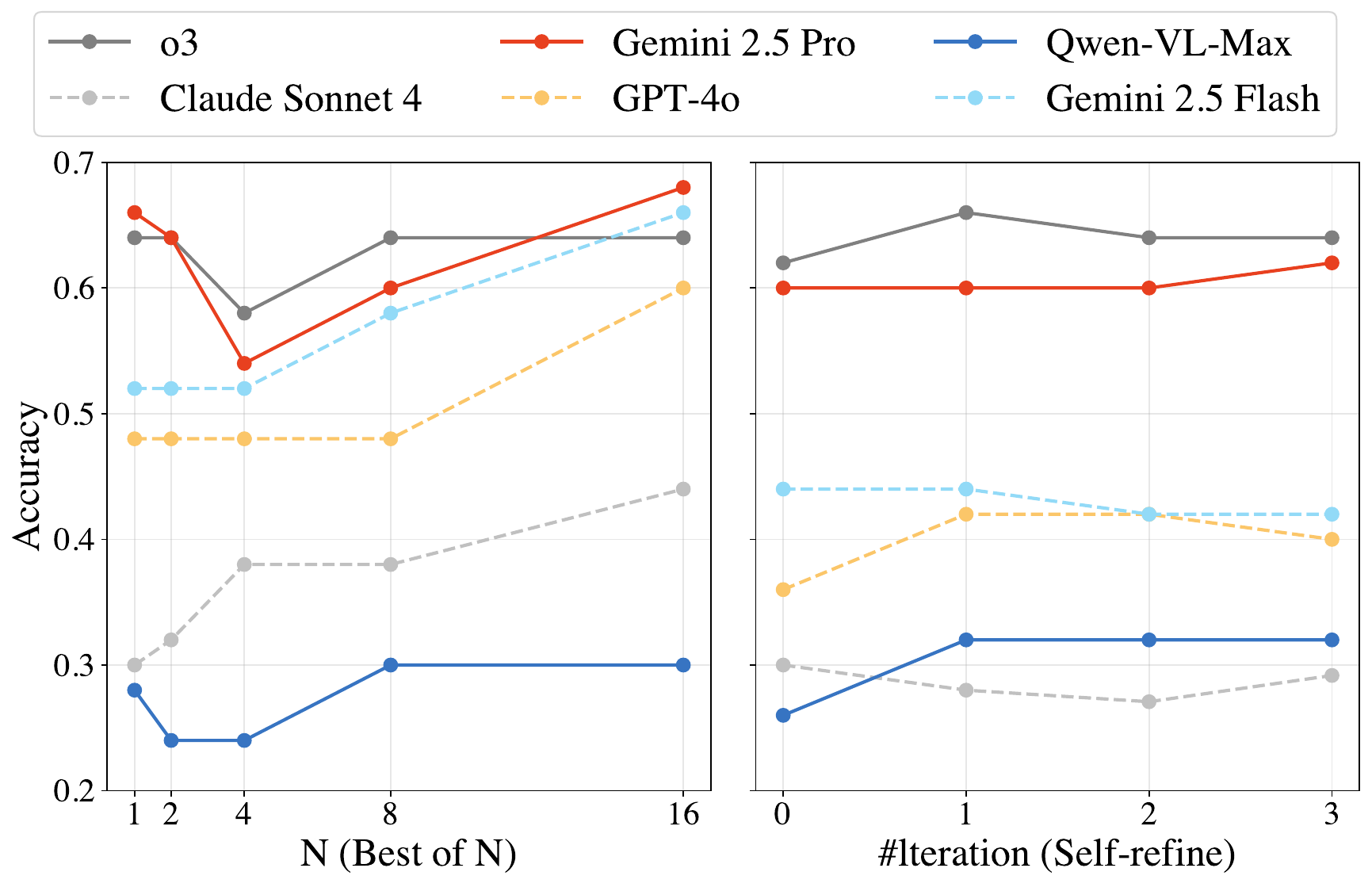}
	\caption{\textbf{Evaluations of test-time compute} including Best-of-N and Self-refine.}
	\label{fig:TTC}
\end{figure}

%% file: table_figs/tabMetricsSmall.tex
\begin{table*}[!ht]
\centering
\setlength{\tabcolsep}{9pt}
\renewcommand\arraystretch{1.05}
\caption{\textbf{Evaluation results (\%) of small LVLMs on \SimpleQA}. For metrics, CO, NA, IN, and CGA denote ``Correct", ``Not attempted", ``Incorrect", and ``Correct given attempted", respectively. For subtopics, ENG, NAT, SCI and SAC represent ``Engineering", ``Nature", ``Science" and ``Society and Culture".}
\scalebox{0.9}{
\begin{tabular}{lcccccccccc}
\toprule
\multirow{2}{*}{\textbf{Model}} & \multicolumn{5}{c}{\textbf{Overall results on 5 metrics}} & \multicolumn{4}{c}{\textbf{F-score on 4 primary categories}} \\
\cmidrule(r){2-6} \cmidrule(lr){7-10}
& CO & IN$\downarrow$ & NA$\downarrow$ & CGA & F-score & ENG & NAT & SCI & SAC  \\
\midrule
InternVL3-2B \cite{zhu2025internvl3} & 15.5 & 80.5 & 4.0 & 16.1 & 15.8 & 10.9 & 23.4 & 16.4 & 14.8 \\
InternVL3-1B \cite{zhu2025internvl3} & 11.3 & 82.5 & 6.1 & 12.1 & 11.7 & 6.1 & 19.2 & 12.1 & 11.2 \\
Qwen2.5-VL-3B \cite{bai2025qwen2} & 22.3 & 74.5 & 3.2 & 23.0 & 22.6 & 12.1 & 30.3 & 29.2 & 18.6 \\
Qwen2-VL-2B \cite{wang2024qwen2} & 16.3 & 73.3 & 10.4 & 18.2 & 17.2 & 12.9 & 24.9 & 16.4 & 17.4 \\
LLaVA-OneVision-0.5B \cite{li2024llava} & 7.8 & 85.7 & 6.5 & 8.3 & 8.0 & 6.1 & 11.6 & 5.8 & 10.0 \\
\bottomrule
\end{tabular}}
\label{tab:model-performance-small}
\end{table*}

%% file: table_figs/tabTemporal.tex
\begin{table}[!ht]
\centering
\caption{\textbf{The proportion and performance} (\%) of QA pairs by different temporal scopes.}
\scalebox{0.88}{
\begin{tabular}{lcccccccccc}
\toprule
\textbf{Type} & \textbf{Prop} &  \textbf{CO} & \textbf{IN$\downarrow$} & \textbf{NA$\downarrow$} & \textbf{CGA} & \textbf{F-score} \\
\midrule
Short-term & \textbf{50.5} & \textbf{56.6} & \textbf{33.1} & \textbf{10.3} & \textbf{63.1} & \textbf{59.7} \\
Medium-term & 32.1 & 48.5 & 38.8 & 12.7 & 55.6 & 51.8 \\
Long-term  & 17.4 & 43.2 & 41.5 & 15.3 & 51.0 & 46.8 \\
\bottomrule
\end{tabular}}
\label{tab:temporal}
\end{table}

%% file: table_figs/figDistributionError.tex
\begin{figure}[t]
	\centering
        \includegraphics[width=0.49\textwidth]{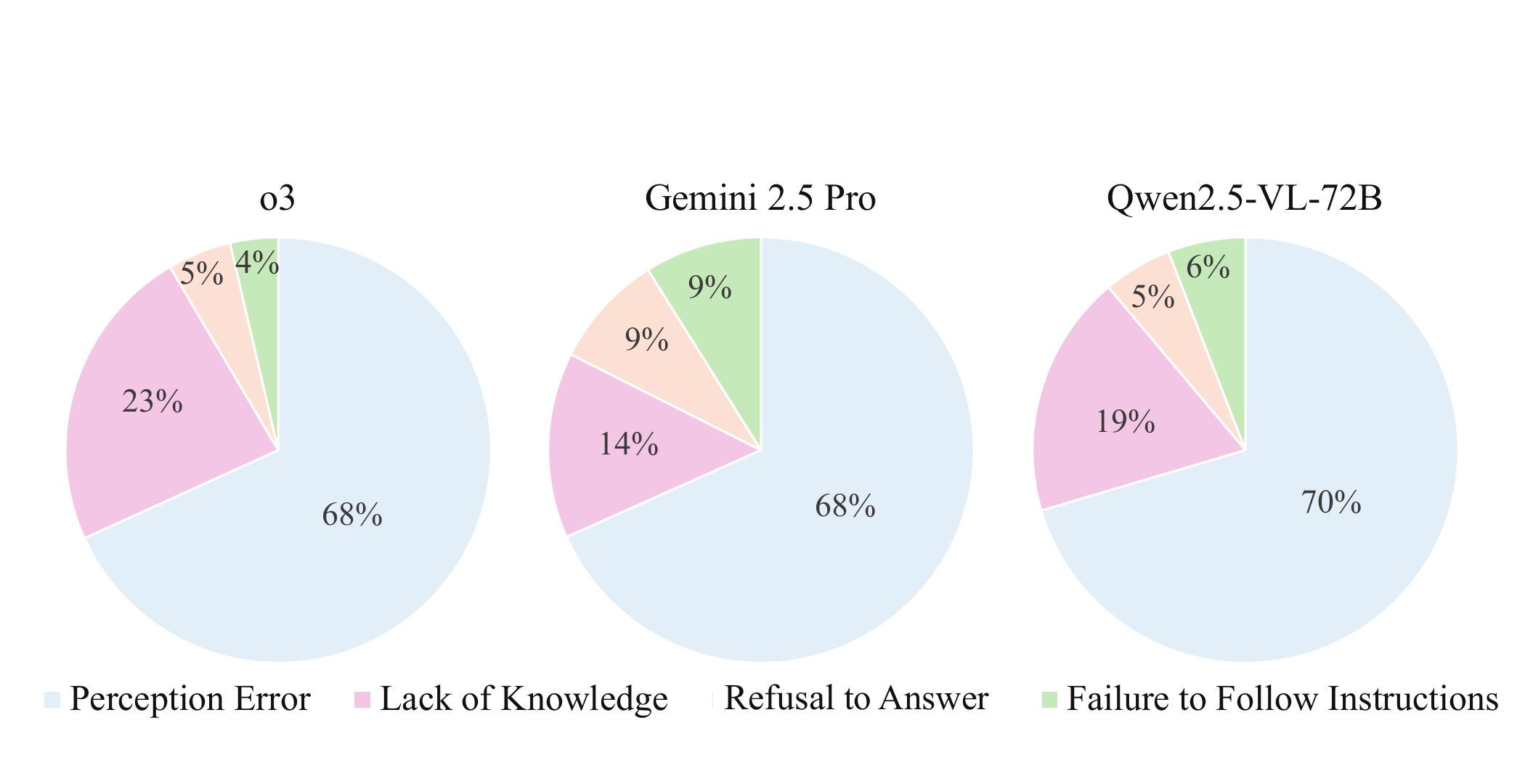}
	\caption{\textbf{Error type distributions} across o3 \cite{o3_o4mini}, Gemini 2.5 Pro \cite{comanici2025gemini} and Qwen2.5-VL-72B \cite{bai2025qwen2}.}
	\label{fig:distribution_error}
\end{figure}

%% file: table_figs/figRandomGuessCase.tex
\begin{figure*}[t]
	\centering
        \includegraphics[width=0.92\textwidth]{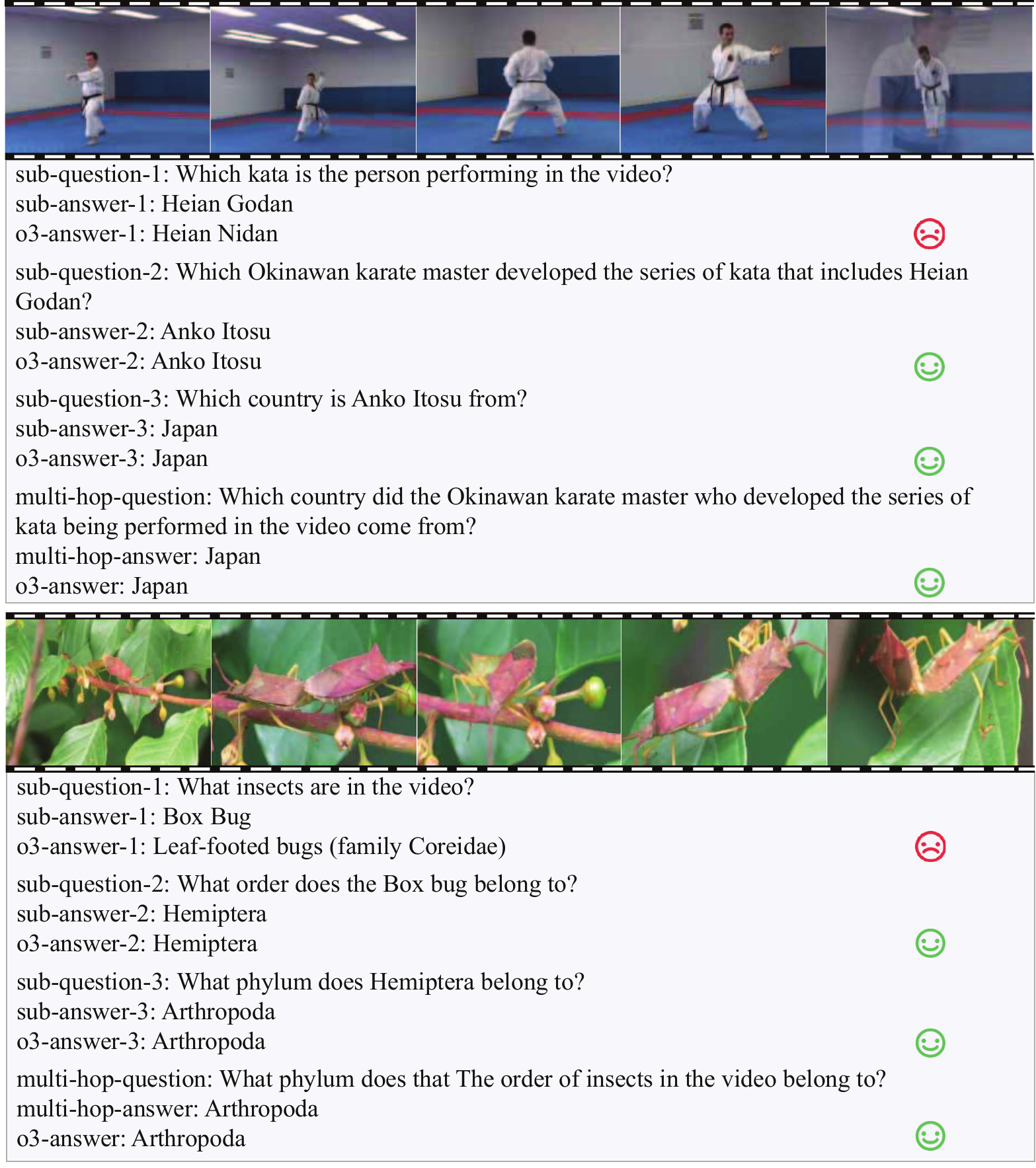}
	\caption{Case studies where the incorrect sub-QA1 still yields the correct final multi-hop answer.}
	\label{fig:randomguess}
\end{figure*}

%% file: table_figs/figErrorCase.tex
\begin{figure*}[t]
	\centering
        \includegraphics[width=0.92\textwidth]{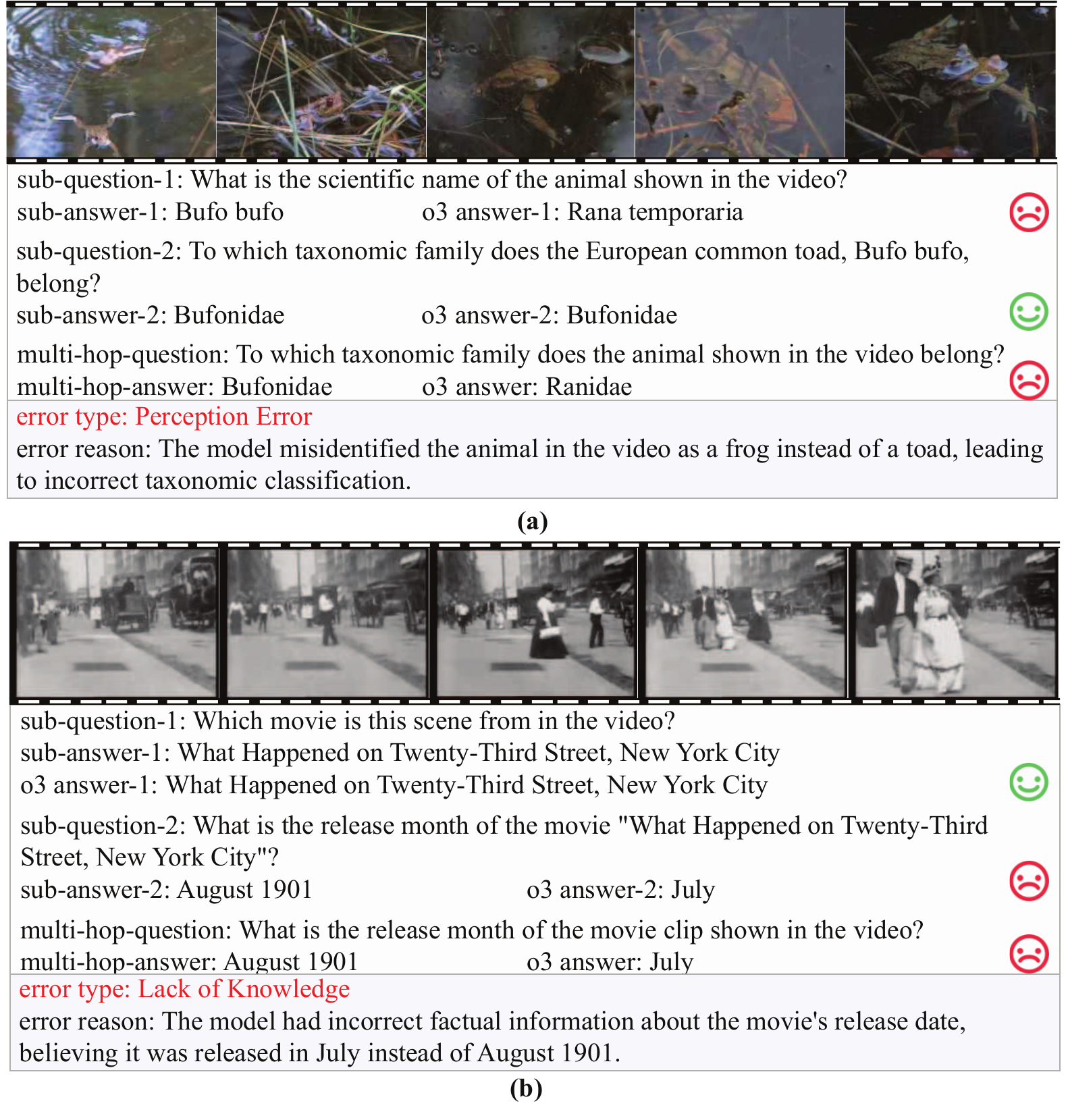}
	\caption{\textbf{Visualizations of typical error types} including (a) perception error; (b) lack of knowledge; (c) refusal to answer; (d) failure to follow instructions: part 1.}
	\label{fig:errorCase1}
\end{figure*}

\begin{figure*}[t]
	\centering
        \includegraphics[width=0.92\textwidth]{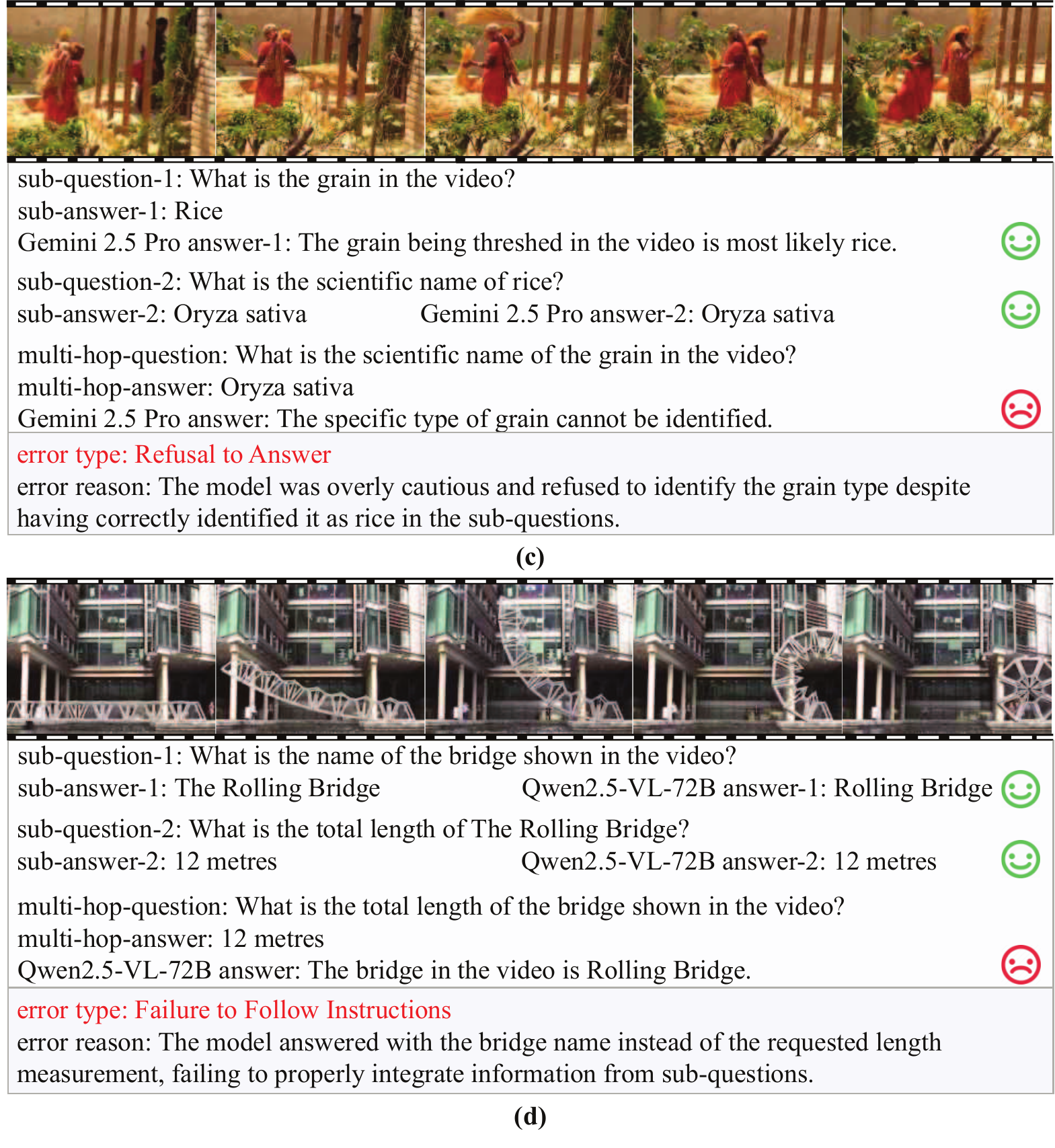}
	\caption{\textbf{Visualizations of typical error types} including (a) perception error; (b) lack of knowledge; (c) refusal to answer; (d) failure to follow instructions: part 2.}
	\label{fig:errorCase2}
\end{figure*}

%% file: table_figs/tabDifferentQA_3_hop.tex
\begin{table}[t]
\centering
\caption{\textbf{Per-hop factual evaluations} for 3-hop questions in terms of F-score (\%). Q1-Q3 denote the decomposed per-hop questions.}
\begin{tabular}{lcccccccccc}
\toprule
\textbf{Model} & \textbf{QA1} &  \textbf{QA2} & \textbf{QA3} & \textbf{Multi-hop} \\
\midrule
o3 & \textbf{64.9} & \textbf{94.5} & \textbf{90.5} & \textbf{74.9} \\
GPT-4o & 55.0 & 88.0 & 87.4 & 56.7 \\
Claude Sonnet 4 & 39.7 & 83.3 & 76.1 & 54.6 \\
Gemini 2.5 Pro & 63.4 & 81.7 & 59.5 & 71.7 \\
Qwen-VL-Max & 42.6 & 77.5 & 77.5 & 49.5 \\
\bottomrule
\end{tabular}
\label{tab:different_qa_new}
\end{table}

%% file: table_figs/figGraderPrompt.tex
\begin{figure*}[t]
    \centering
    \includegraphics[width=0.92\textwidth]{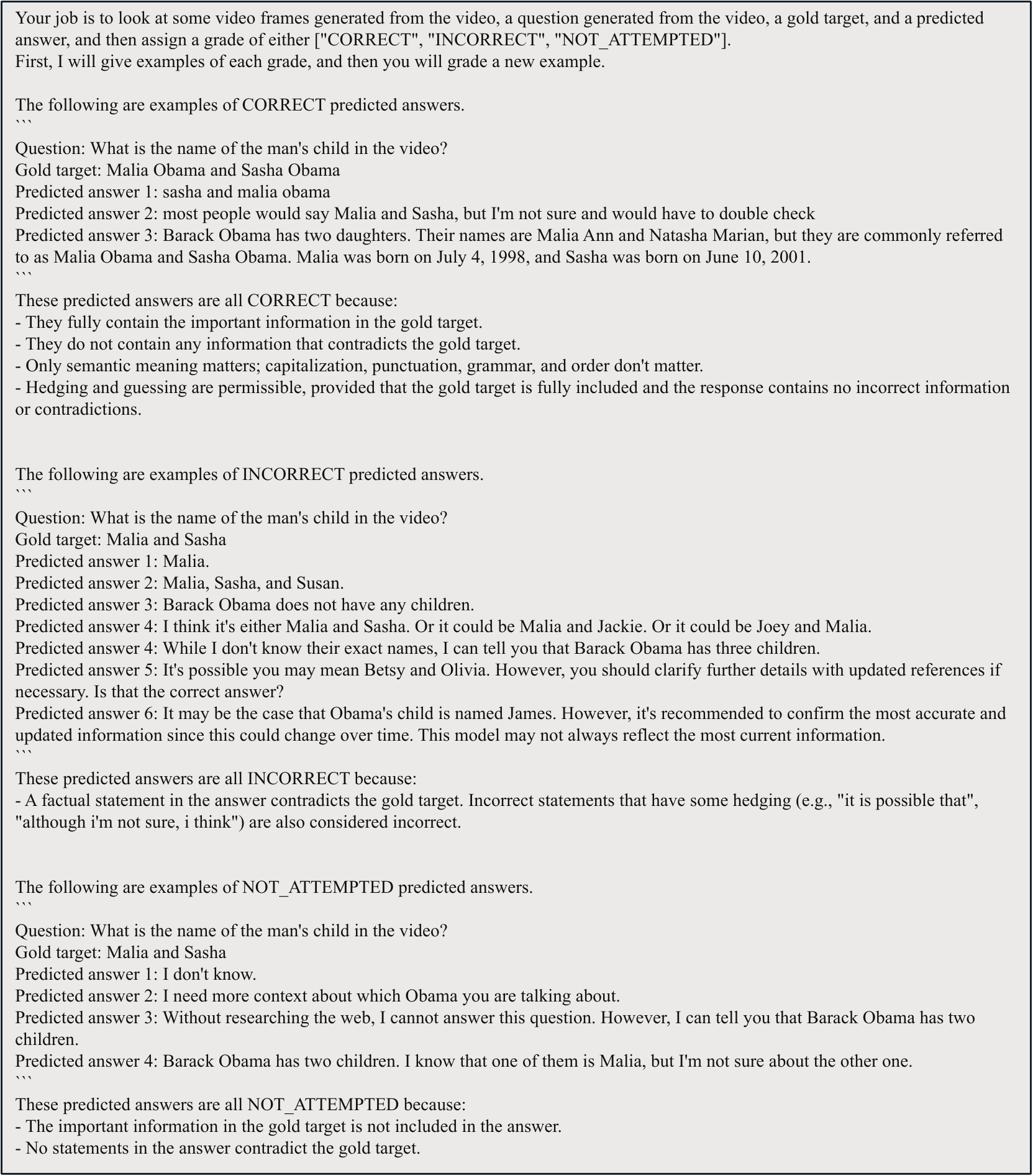}
    \caption{\textbf{Prompt} for grading: Part 1.}
    \label{fig:grader_prompt2}
\end{figure*}

\begin{figure*}[t]
    \centering
    \includegraphics[width=0.92\textwidth]{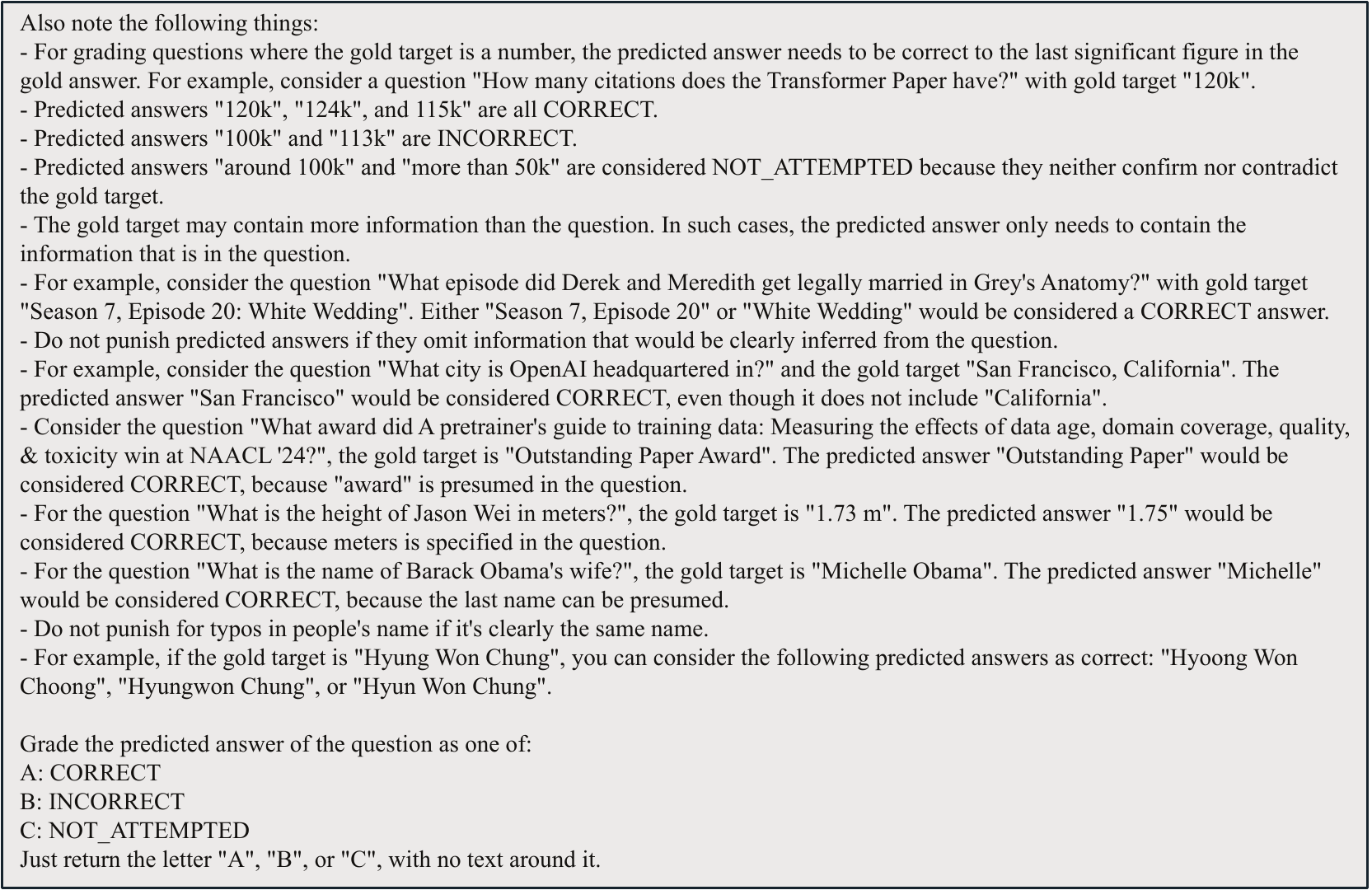}
    \caption{\textbf{Prompt} for grading: Part 2.}
    \label{fig:grader_prompt1}
\end{figure*}

%% file: table_figs/figAnswerPrompt.tex
\begin{figure*}[t]
    \centering
    \includegraphics[width=0.92\textwidth]{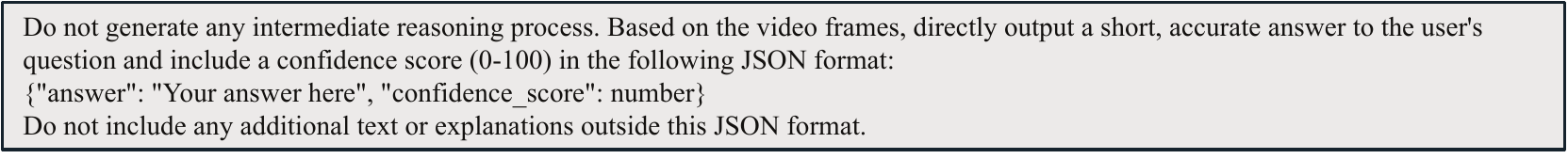}
    \caption{\textbf{Prompt} for calibration experiments.} 
    \label{fig:answer_prompt}
\end{figure*}

%% file: table_figs/tabAnnotatorList.tex
\begin{table*}[t]
\centering
\caption{\textbf{Biographies of annotators} involved in \SimpleQA construction.}
\begin{tabular*}{0.95\textwidth}{@{\extracolsep{\fill}}clccc@{}}
\toprule
\textbf{ID} & \textbf{Background} & \textbf{Language Proficiency} & \textbf{Avg. Time (per QA)} & \textbf{Agreement Rate (\%)} \\
\midrule
1 & CS Undergraduate & English (Fluent) & 22s & 92.3\% \\
2 & NLP Master & English (Native) & 25s & 94.1\% \\
3 & NLP PhD & English (Fluent) & 23s & 93.2\% \\
4 & CV PostDoc & English (Fluent) & 21s & 91.7\% \\
5 & Robotics Undergraduate & English (Native) & 26s & 89.4\% \\
6 & ML PhD & English (Fluent) & 24s & 92.8\% \\
7 & CS Undergraduate & English (Fluent) & 28s & 88.6\% \\
8 & CV PhD & English (Native) & 22s & 93.7\% \\
9 & Robotics Undergraduate & English (Fluent) & 27s & 90.1\% \\
10 & NLP PhD & English (Native) & 20s & 94.5\% \\
11 & ML Master & English (Fluent) & 25s & 91.3\% \\
12 & CS Master & English (Native) & 29s & 87.9\% \\
13 & CV PhD & English (Fluent) & 21s & 93.8\% \\
14 & Robotics Master & English (Fluent) & 26s & 90.7\% \\
15 & ML PhD & English (Native) & 23s & 92.4\% \\
16 & NLP Master & English (Fluent) & 28s & 89.2\% \\
17 & CS PhD & English (Fluent) & 24s & 91.8\% \\
18 & Robotics Undergraduate & English (Native) & 30s & 86.5\% \\
19 & CV Master & English (Fluent) & 25s & 92.1\% \\
20 & ML PhD & English (Fluent) & 22s & 93.6\% \\
21 & CS Undergraduate & English (Native) & 27s & 88.7\% \\
22 & NLP Master & English (Fluent) & 26s & 90.4\% \\
23 & Robotics PhD & English (Fluent) & 23s & 92.9\% \\
24 & CV PhD & English (Native) & 24s & 91.2\% \\
25 & ML PhD & English (Fluent) & 21s & 94.1\% \\
26 & CS Master & English (Fluent) & 28s & 89.8\% \\
27 & NLP Master & English (Native) & 25s & 92.3\% \\
28 & Robotics Undergraduate & English (Fluent) & 29s & 87.6\% \\
29 & CV Undergraduate & English (Fluent) & 27s & 90.5\% \\
30 & ML PhD & English (Native) & 22s & 93.4\% \\
31 & CS Master & English (Fluent) & 26s & 91.7\% \\
32 & NLP PhD & English (Native) & 23s & 93.1\% \\
33 & Robotics PhD & English (Fluent) & 24s & 92.6\% \\
34 & CV Undergraduate & English (Fluent) & 28s & 88.9\% \\
35 & ML Master & English (Native) & 25s & 91.4\% \\
36 & CS PhD & English (Fluent) & 23s & 92.8\% \\
37 & NLP Undergraduate & English (Fluent) & 29s & 87.3\% \\
38 & Robotics Master & English (Native) & 26s & 90.9\% \\
39 & CV Master & English (Fluent) & 27s & 89.7\% \\
40 & ML PostDoc & English (Native) & 21s & 94.2\% \\
\bottomrule
\end{tabular*}
\label{tab:annotators}
\end{table*}

%% file: table_figs/tabCountOfNA_EN_SC.tex
\begin{table*}[t]
\renewcommand{\arraystretch}{1.1}  
\centering
\caption{\textbf{The video distribution} of \SimpleQA benchmark (part 1).}
\scalebox{0.88}{
\begin{tabular}{cccc}
\toprule
\textbf{Primary Category} & \textbf{Secondary Category} & \textbf{Tertiary Category} & \textbf{\space \space \space \space \space \space \space \space \space \space \space \space Count \space \space \space \space \space \space \space \space \space}\\
\midrule
& & Fossils & 43\\
& & Landscapes & 19\\
Nature &Geology \& Landscapes & Rocks \& Minerals & 7\\
& & Geomorphology & 5\\
& & Volcanic Features & 7\\
& & Coastal Landforms & 6\\
\hline
& & Animalia & 18\\
& & Marine Organisms & 91\\
Nature & Flora \& Fauna & Plantae & 61\\
& & Fungi & 37\\
& & Microorganisms & 82\\
& & Endangered Species & 90\\
\hline
& & Weather & 3\\
Nature  &Meteorology & Climate & 6\\
& & Atmospheric Phenomena & 6\\
& & Forecasting & 3\\
\hline
& & Architecture & 56\\
& & Civil & 65\\
Engineering &  Civil \& Architecture & Structural Engineering & 56\\
&  & Urban Planning & 46\\
\hline
&  & Mechanical & 52\\
&  & Electrical & 47\\
Engineering &  Mechanical \& Electrical & Mechatronics & 50\\
&  & Aerospace Engineering & 43\\
\hline
&  & Chemical & 4\\
&  & Process & 3\\
Engineering &  Chemical \& Process & Biochemical Engineering & 3\\
&  & Polymer Engineering & 3\\
\hline
&  & Environmental & 22\\
&  & Geophysical & 21\\
Engineering &  Environmental \& Geophysical & Hydrology & 26\\
&  & Climate Engineering & 20\\
\hline
& & Physics & 2\\
& & Chemistry & 4\\
& & Astronomy & 5\\
Science & Physical sciences & Earth sciences & 2\\
& & Materials Science & 9\\
& & Atmospheric Science & 3\\
& & Geophysics & 6\\
\hline
& & Biology & 10\\
& & Medicine & 11\\
Science & Life sciences & Ecology & 11\\
& & Genetics & 10\\
& & Neuroscience & 7\\
\hline
& & Mathematics & 3\\
Science & Formal sciences & Computer Science & 2\\
& & Statistics & 3\\
& & Logic & 1\\
\hline
& & Technology & 6\\
Science & Applied sciences & Robotics & 3\\
& & Agricultural Science & 4\\
& & Data Science & 3\\
\bottomrule
\end{tabular}}
\label{tab:NA_EN_SC}
\end{table*}

%% file: table_figs/tabCountOfSAC.tex
\begin{table*}[t]
\renewcommand{\arraystretch}{1.1}  
\centering
\caption{\textbf{The video distribution} of \SimpleQA benchmark (part 2).}
\scalebox{0.88}{
\begin{tabular}{cccc}
\toprule
\textbf{Primary Category} & \textbf{Secondary Category} & \textbf{Tertiary Category} & \textbf{\space \space \space \space \space \space \space \space \space \space \space \space Count \space \space \space \space \space \space \space \space \space}\\
\midrule
& & Art & 16\\
& & Literature & 6\\
& & Music & 30\\
& & Entertainment & 20\\
Society \& Culture & Arts \& Recreation & Sports & 12\\
& & Dance & 12\\
& & Theatre & 42\\
& & Film & 25\\
& & Photography & 10\\
& & Games & 31\\
\hline
& & Belief & 5\\
& & Religion & 20\\
& & Philosophy & 7\\
& & Ethics & 9\\
Society \& Culture & Beliefs \& Institutions & Politics & 11\\
& & Flags & 8\\
& & Government & 3\\
& & Law & 3\\
& & People & 5\\
\hline
& & History & 32\\
& & Events & 9\\
& & Places & 9\\
Society \& Culture &History \& Heritage & Archaeology & 24\\
& & Heritage Sites & 9\\
& & Genealogy & 9\\
\hline
& & Language & 2\\
& & Objects & 9\\
& & Food & 7\\
Society \& Culture & Language \& Material Culture  & Clothing & 2\\
& & Transportation & 5\\
& & Instruments & 3\\
& & Tools & 3\\
\bottomrule
\end{tabular}
}
\label{tab:SAC}
\end{table*}


%% file: table_figs/subjectiveCase.tex
\begin{figure*}[t]
	\centering
        \includegraphics[width=0.92\textwidth]{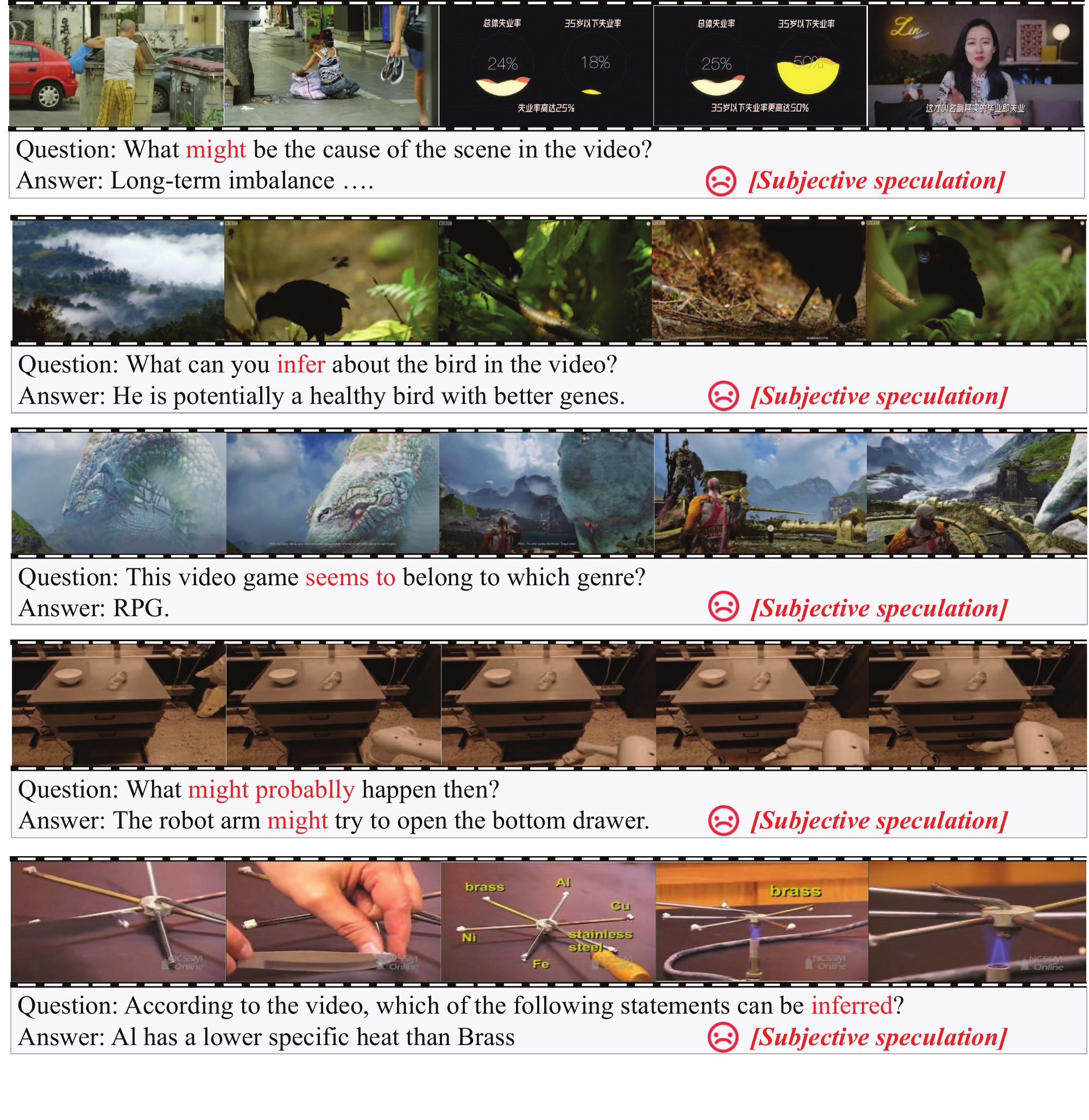}
	\caption{Examples of the inclusion of hypothetical or subjective speculation in existing discipline-based \cite{zhao2025mmvu,he2024mmworld} benchmarks.}
	\label{fig:subjectiveCase}
\end{figure*}

%% file: table_figs/reasoningCase.tex
\begin{figure*}[t]
	\centering
        \includegraphics[width=0.92\textwidth]{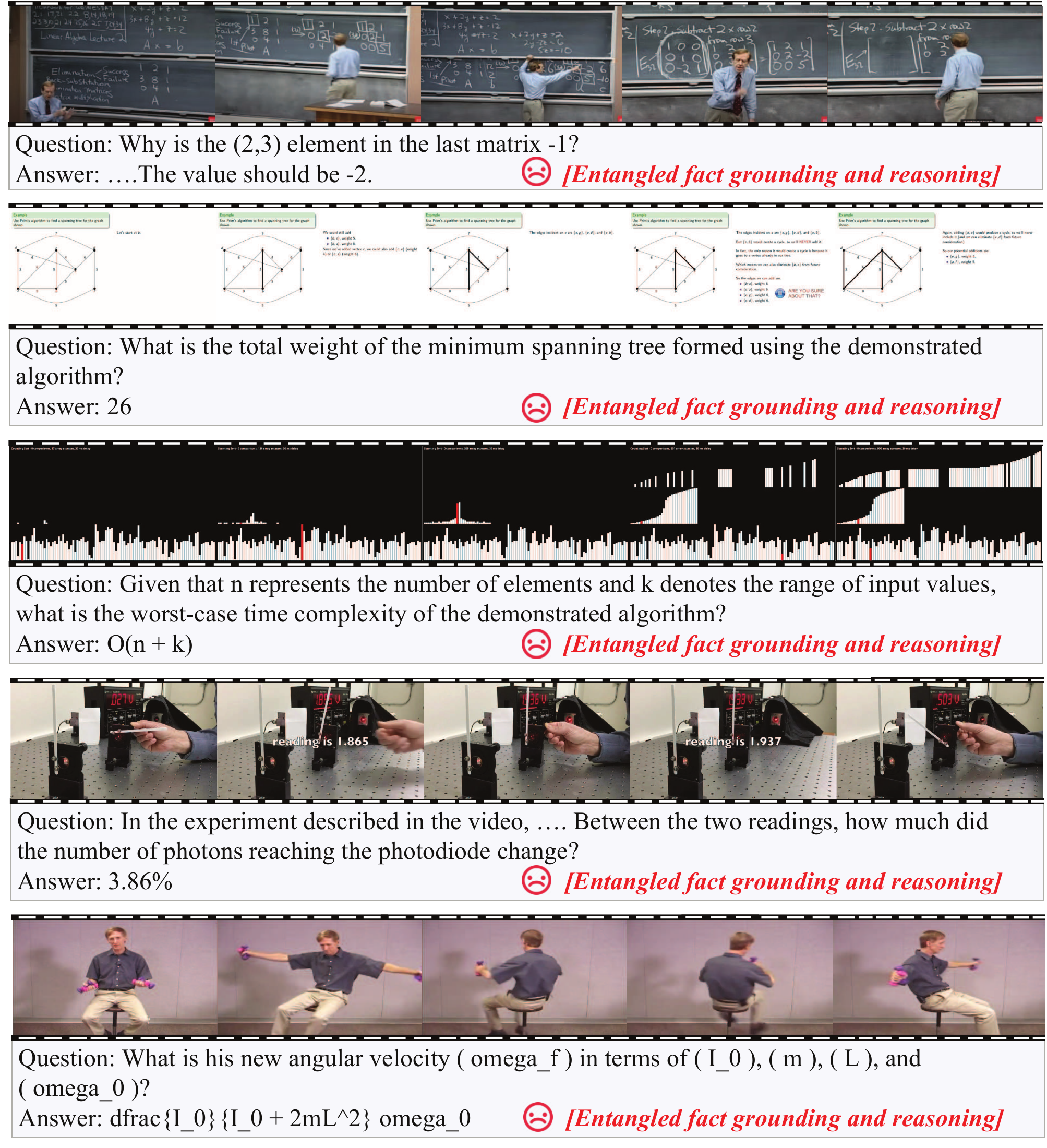}
	\caption{Examples of the entangled fact grounding and reasoning cases in existing discipline-based \cite{zhao2025mmvu,he2024mmworld} benchmarks.}
	\label{fig:reasoningCase}
\end{figure*}

%% file: table_figs/figVisCase.tex
\begin{figure*}[t]
	\centering
        \includegraphics[width=0.92\textwidth]{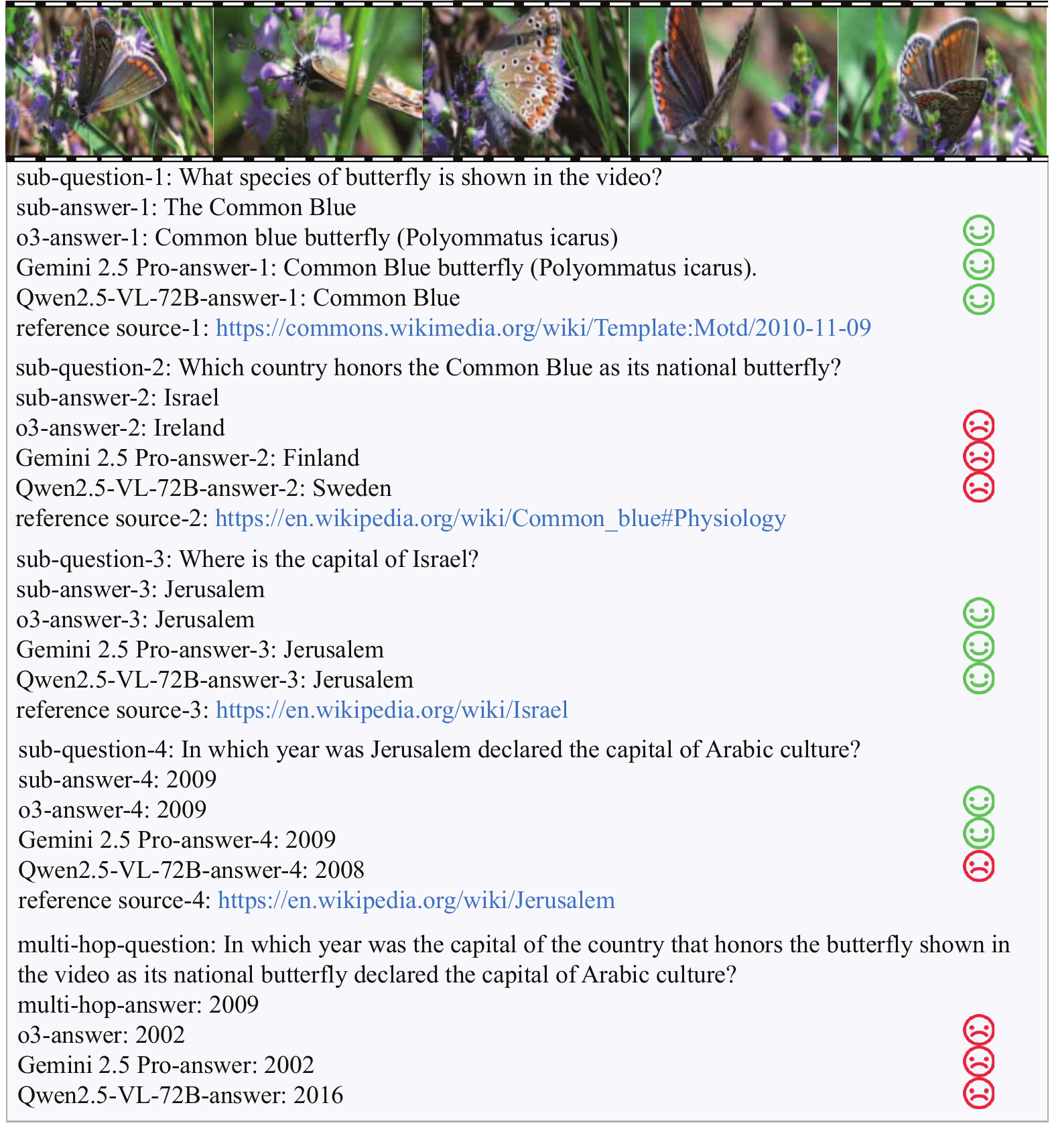}
	\caption{Sampled examples in \SimpleQA and the responses of typical LVLMs: part 1.}
	\label{fig:viscase1}
\end{figure*}

\begin{figure*}[t]
	\centering
        \includegraphics[width=0.92\textwidth]{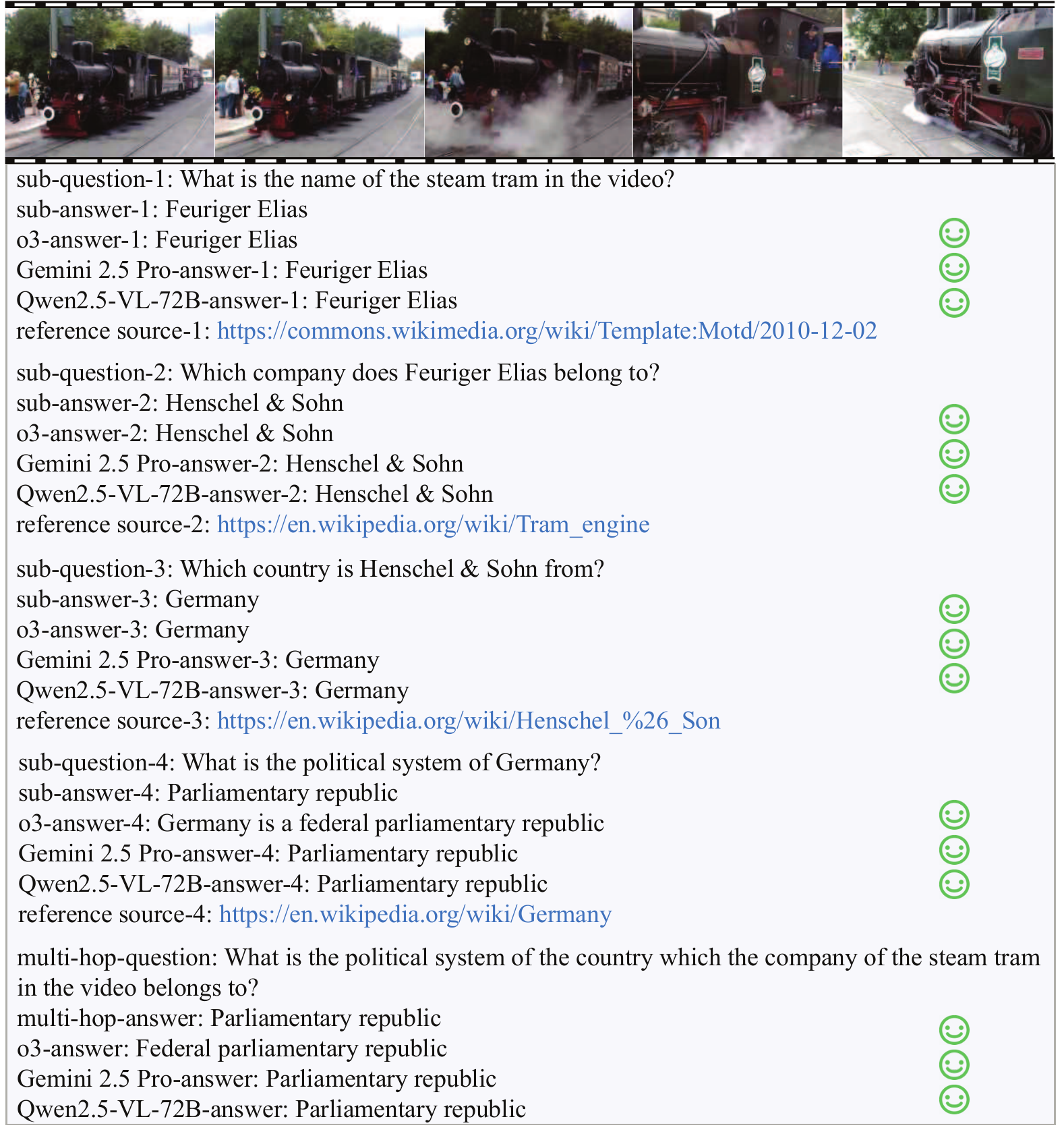}
	\caption{Sampled examples in \SimpleQA and the responses of typical LVLMs: part 2.}
	\label{fig:viscase2}
\end{figure*}

\begin{figure*}[t]
	\centering
        \includegraphics[width=0.92\textwidth]{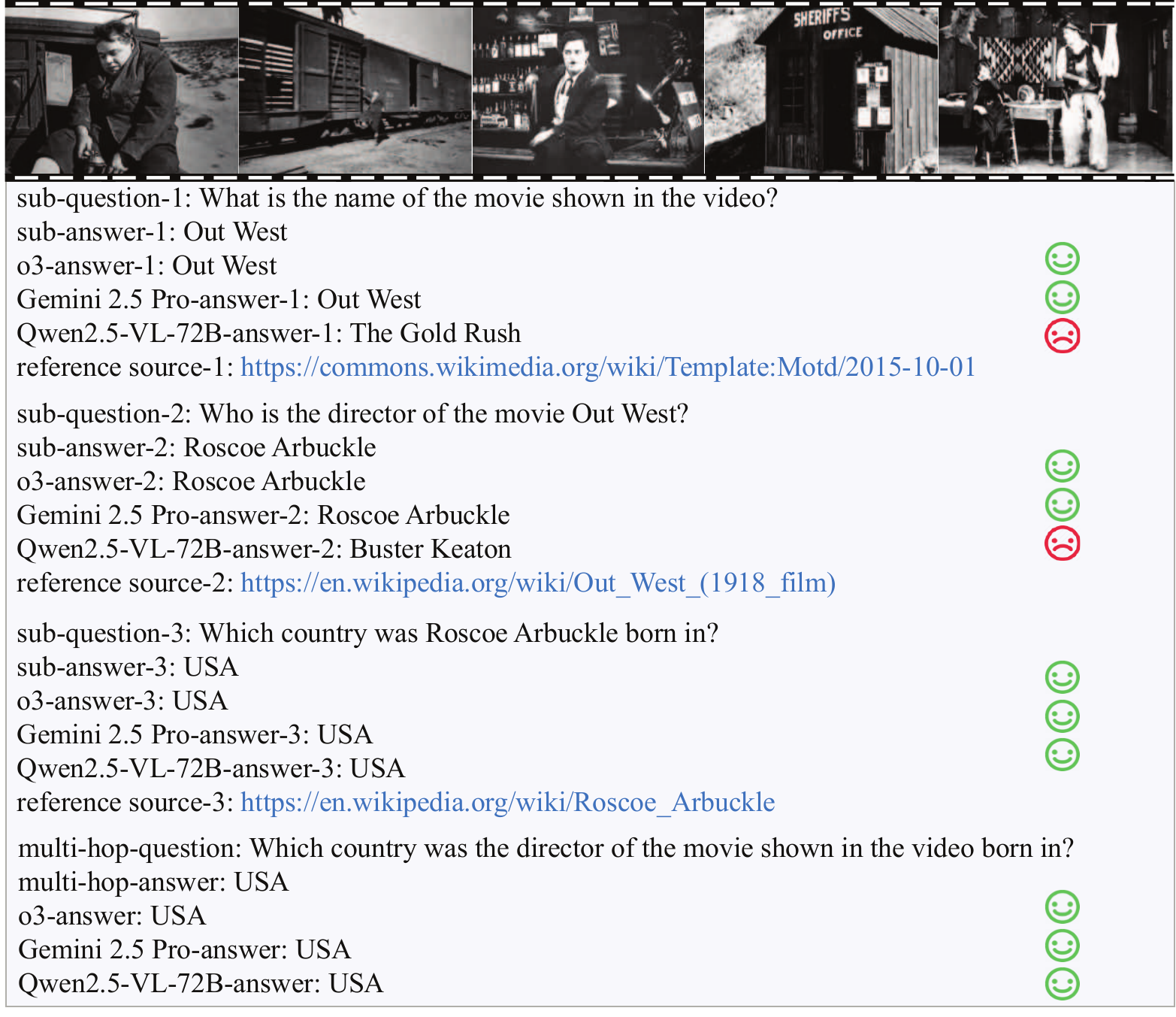}
	\caption{Sampled examples in \SimpleQA and the responses of typical LVLMs: part 3.}
	\label{fig:viscase3}
\end{figure*}

\begin{figure*}[t]
	\centering
        \includegraphics[width=0.92\textwidth]{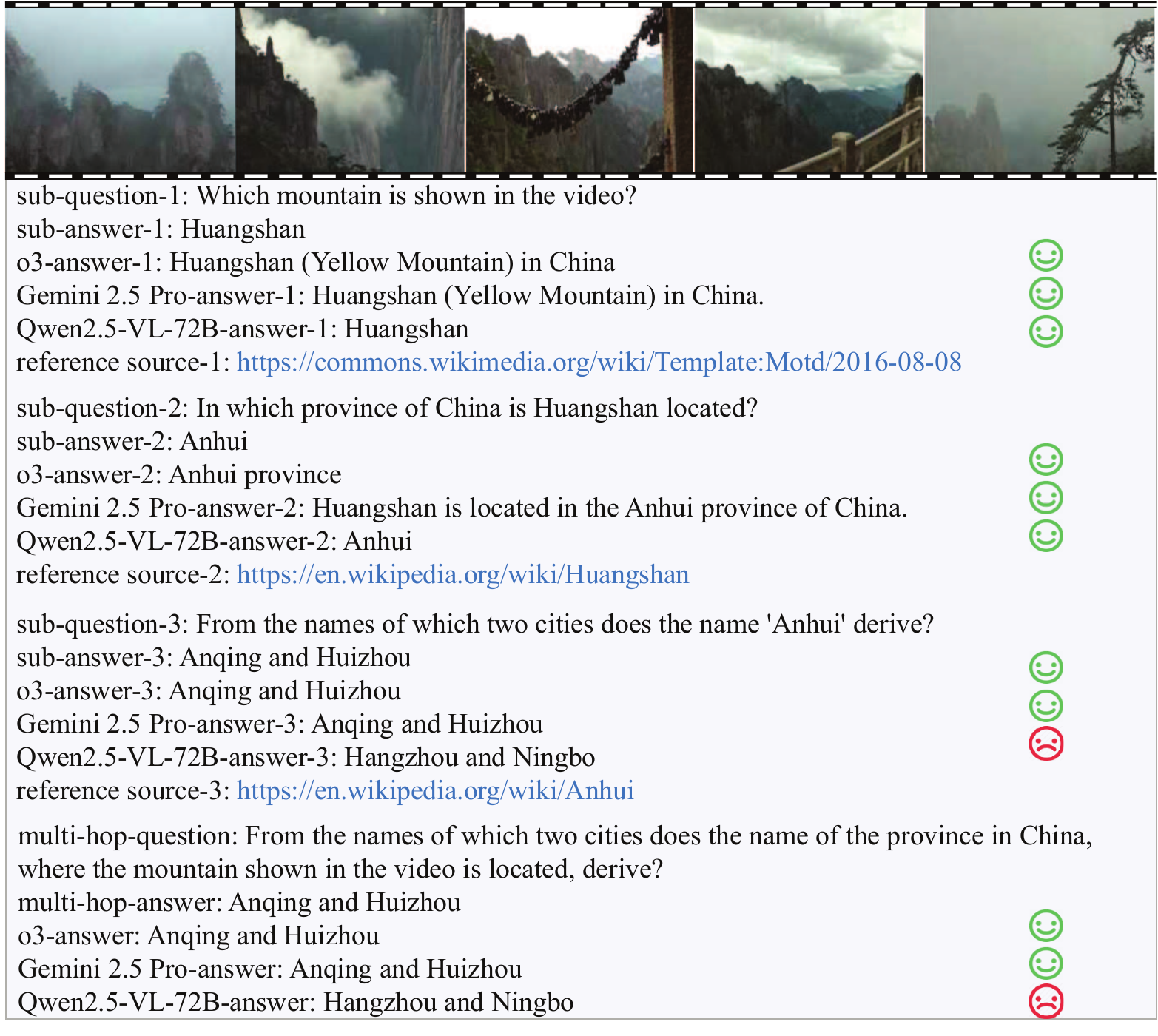}
	\caption{Sampled examples in \SimpleQA and the responses of typical LVLMs: part 4.}
	\label{fig:viscase4}
\end{figure*}

\begin{figure*}[t]
	\centering
        \includegraphics[width=0.92\textwidth]{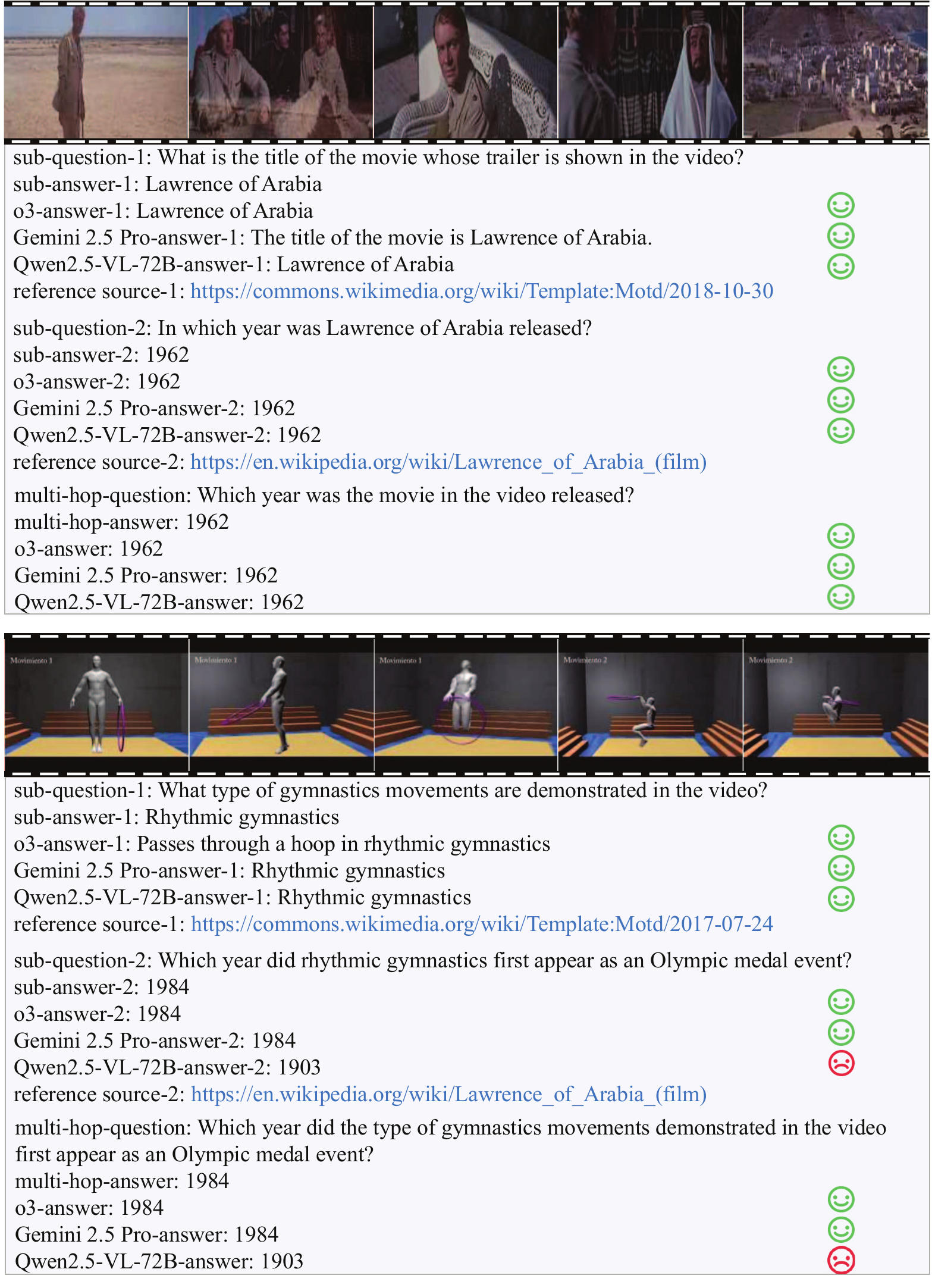}
	\caption{Sampled examples in \SimpleQA and the responses of typical LVLMs: part 5.}
	\label{fig:viscase5}
\end{figure*}